\documentclass{article}
\usepackage{natbib}
\setcitestyle{square,numbers}
\usepackage[final]{neurips_2022}
\usepackage{amsmath}
\usepackage{amssymb}
\usepackage{mathtools}
\usepackage{amsthm}

\usepackage{graphicx}

\usepackage{makecell}

\usepackage[utf8]{inputenc} 
\usepackage[T1]{fontenc}    
\usepackage{url}            
\usepackage{booktabs}       
\usepackage{amsfonts}       
\usepackage{nicefrac}       
\usepackage{microtype}      
\usepackage{xcolor}         
\usepackage{algorithm}
\usepackage{algorithmic}
\usepackage{amsmath,amssymb}
\usepackage{mathtools}
\usepackage[colorlinks=true,bookmarks=false]{hyperref}
\usepackage{bm}
\usepackage{subcaption}
\usepackage{comment}
\usepackage{wrapfig}
\usepackage[capitalize,noabbrev]{cleveref}

\theoremstyle{plain}

\theoremstyle{definition}

\theoremstyle{remark}

\usepackage[textsize=tiny]{todonotes}

\providecommand{\argmax}{\operatornamewithlimits{argmax}} 
\providecommand{\argmin}{\operatornamewithlimits{argmin}} 
\providecommand{\R}{\mathbb{R}} 
\providecommand{\E}{\mathbb{E}} 
\renewcommand{\geq}{\geqslant} 
\renewcommand{\leq}{\leqslant} 
\DeclarePairedDelimiterX{\inner}[2]{\langle}{\rangle}{#1, #2}

\DeclarePairedDelimiter{\abs}{\lvert}{\rvert}

\usepackage{color}

\renewcommand{\paragraph}[1]{\par\noindent\textbf{#1}\quad}

\providecommand{\proposed}{M2TD3} 

\title{Max-Min Off-Policy Actor-Critic Method Focusing on Worst-Case Robustness to Model Misspecification}

\author{%
  Takumi Tanabe \\
  University of Tsukuba \& RIKEN AIP\\
  Tsukuba, Ibaraki 305-8573, Japan \\
  \texttt{tanabe@bbo.cs.tsukuba.ac.jp} \\
  \And
  Rei Sato \\
  University of Tsukuba \& RIKEN AIP\\
  Tsukuba, Ibaraki 305-8573, Japan \\
  \texttt{reisato@bbo.cs.tsukuba.ac.jp} \\
  \And
  Kazuto Fukuchi \\
  University of Tsukuba \& RIKEN AIP\\
  Tsukuba, Ibaraki 305-8573, Japan \\
  \texttt{fukuchi@cs.tsukuba.ac.jp} \\
  \And
  Jun Sakuma \\
  University of Tsukuba \& RIKEN AIP\\
  Tsukuba, Ibaraki 305-8573, Japan \\
  \texttt{jun@cs.tsukuba.ac.jp} \\
  \And
  Youhei Akimoto \\
  University of Tsukuba \& RIKEN AIP\\
  Tsukuba, Ibaraki 305-8573, Japan \\
  \texttt{akimoto@cs.tsukuba.ac.jp} \\
}

\begin{document}

\maketitle

\begin{abstract}
In the field of reinforcement learning, because of the high cost and risk of policy training in the real world, policies are trained in a simulation environment and transffered to the corresponding real-world environment.
However, differences in the environments lead to model misspecification.
Multiple studies report significant deterioration of policy performance in a real-world environment.
In this study, we focus on scenarios involving a simulation environment with uncertainty parameters and the set of their possible values, called the uncertainty parameter set.
The aim is to optimize the worst-case performance on the uncertainty parameter set to guarantee the performance in the corresponding real-world environment.
To obtain a policy for the optimization, we propose an off-policy actor-critic approach called the Max-Min Twin Delayed Deep Deterministic Policy Gradient algorithm (\proposed{}), which solves a max-min optimization problem using a simultaneous gradient ascent descent approach.
Experiments in multi-joint dynamics with contact (MuJoCo) environments show that the proposed method exhibited a worst-case performance superior to several baseline approaches.
Our implementation is publicly available (\url{https://github.com/akimotolab/M2TD3}).
\end{abstract}

\section{Introduction}

Applications of deep reinforcement learning (DRL) to control tasks that involving interaction with the real world are limited because of the need for interaction between the agent and the real-world environment~\cite{tobin2017domain}.
This difficulty arises because of high interaction time, agent maintenance costs, and safety issues during the interaction.
Without enough interactions, DRL tends to overfit to specific interaction histories and yields policies with poor generalizability and safety issues~\cite{peng2018sim}.

To solve this problem, simulation environments that estimate the characteristics of real-world environments are often employed, which enable enough interactions for effective DRL application.
The policy is trained in the simulation environment and then transferred to the real-world environment~\cite{DBLP:conf/rss/YuTLT17}.
Moreover, because fast and general-purpose simulators, such as Open Dynamics Engine~\cite{smith2005open}, Bullet~\cite{coumans2013bullet} and multi-joint dynamics with contact (MuJoCo~~\cite{todorov2012mujoco}), are available in the field of robotics, the cost of developing a simulation environment can sometimes be significantly lower than that for real-world interactions.

However, because a simulation environment is an estimation of the real-world environment, discrepancies exist between the two~\cite{ajay2018augmenting}.
Even if they are entirely similar, discrepancies may still occur, e.g., because of robot degradation over time or weight changes because of part replacements~\cite{Mankowitz2020Robust}.
Studies report significant degradation of policy trained in a simulation environment when transferred to the corresponding real-world environment~\cite{golemo2018sim}.
The above discrepancies limit the use of simulator environments in practice and, thus, limit the use of DRL.

In this study, to guarantee performance in the corresponding real-world environment, we aim to obtain a policy that maximizes the expected reward under the worst-case scenario in the uncertainty parameter set $\Omega$.
We focus on a scenario where (1) a simulator $\mathcal{M}_\omega$ is available, (2) the model uncertainty is parameterized by $\omega \in \Omega$, and $\omega$ is configurable in the simulation, and (3) the real-world environment is identified with some $\omega^* \in \Omega$, where $\omega^*$ is fixed during an episode; however, (4) $\omega^*$ is unknown, is not uniquely determined, or changes from episode to episode\footnote{Here are two examples where hypotheses~(3) and (4) are reasonable. (I) We train a common controller of mass-produced robot manipulators that are slightly different due to errors in the manufacturing process ($\omega^*$ is fixed for each product, but varies in $\Omega$ from product to product). (II) We train a controller of a robot manipulator whose dynamics change over time because of aging, but each episode is short enough that the change in dynamics during an episode is negligible.}.
The model of uncertainty described in hypotheses~(3) and (4) is referred to as the stationary uncertainty model, in contrast to the time-varying uncertainty model where $\omega^*$ can vary during an episode \cite{DBLP:journals/ior/NilimG05}.
We develop an off-policy deep deterministic actor-critic approach to optimize the worst-case performance, called the Max-Min Twin Delayed Deep Deterministic Policy Gradient Algorithm (\proposed{}).
By extending the bi-level optimization formulation of the standard actor-critic method~\cite{DBLP:journals/corr/PfauV16}, we formulate our problem as a tri-level optimization, where the critic network models a value for each tuple of state, action, and uncertainty parameter, and the actor network models the policy of an agent and an estimate of the worst uncertainty parameter value.
Based on the existing deep deterministic actor-critic approach~\cite{DBLP:journals/corr/LillicrapHPHETS15,fujimoto2018addressing}, we design an algorithm to solve the tri-level optimization problem by implementing several technical components to stabilize the training, including multiple uncertainty parameters, their refreshing strategies, the uncertainty parameter sampler for interaction, and the soft-min policy update.
Numerical experiments on 19 MuJoCo tasks reveal the competitive and superior worst-case performance of our proposed approaches compared to those of several baseline approaches.

\section{Preliminaries}\label{sec:preliminaries}
We consider an episodic Markov Decision Process (MDP)~\cite{domingues2021episodic} family $\mathcal{M}_{\Omega} = \{\mathcal{M}_{\omega}\}_{\omega \in \Omega}$, where $\mathcal{M}_{\omega} = \langle S, A, p_\omega, p^0_{\omega}, r_\omega, \gamma\rangle$ is the MDP with uncertainty parameter $\omega \in \Omega$. The state space $S$ and the action space $A$ are subsets of real-valued vector spaces. The transition probability density $p_\omega: S \times A \times S \to \R$, the initial state probability density $p^0_\omega: S \to \R$, and the immediate reward $r_\omega: S \times A \to \R$ depend on $\omega$. The discount factor is denoted by $\gamma \in (0, 1)$.

We focus on an episodic task on $\mathcal{M}_\Omega$. Let $\mu_{\theta}: S \to A$ be a deterministic policy parameterized by $\theta \in \Theta$. Given an uncertainty parameter $\omega \in \Omega$, the initial state follows $s_0 \sim p_\omega^0$. At each time step $t \geq 0$, the agent observes state $s_t$, select action $a_t = \mu_{\theta}(s_t)$, interacts with the environment, and observes the next state $s_{t}' \sim p_\omega(\cdot \mid s_t, a_t)$, the immediate reward $r_{t} = r_\omega(s_t, a_t)$, and the termination flag $h_t$. Here, $h_t = 1$ if $s_t'$ is a terminal state or a predefined maximal time step for each past episode; otherwise, $h_t = 0$.
If $h_{t} = 0$, let $s_{t+1} = s_{t}'$; otherwise, the next state is reset by $s_{t+1} \sim p_{\omega}^{0}$.
For simplicity, we let $q_\omega(s_{t+1}\mid s_t, a_t)$ be the probability density of $s_{t+1}$ given $s_{t}$ and $a_t$. The discount reward of the trajectory starting from time step $t$ is $R_t = \sum_{k \geq 0} \gamma^{k} r_{t+k}$.

The action value function $Q^{\mu_\theta}(s, a, \omega)$ under $\omega$ is the expectation of $R_t$ starting with $s_t = s$ and $a_t = a$ under $\omega$; that is,
$Q^{\mu_\theta}(s, a, \omega) = \E[R_t \mid s_t = s, a_t = a,
    s_{t+k+1} \sim q_\omega(\cdot\mid s_{t+k}, \mu_{\theta}(s_{t+k}))\ \forall k \geq 0 ]$,
where the expectation is taken over $s_{t+k+1}$ for $k \geq 0$.
Note that we introduce $\omega$ to the argument to explain the $Q$-value dependence on $\omega$; however, this is essentially the same definition as that for the standard action value function $Q^{\mu_\theta}(s, a)$.
The recursive formula is as follows:
\begin{equation}
    Q^{\mu_\theta}(s, a, \omega) = r_\omega(s, a) + \gamma\E[Q^{\mu_\theta}(s', \mu_\theta(s'), \omega) \mid s' \sim q_\omega(\cdot\mid s, a)]\enspace.\label{eq:q-recursive}
\end{equation}

We consider an off-policy reinforcement learning setting, where the agent interacts with $\mathcal{M}_\omega$ using a stochastic behavior policy $\beta: S \times A \to \R$.
However, the target policy to be optimized, denoted by $\mu_{\theta}$, is deterministic.
Under a fixed $\omega \in \Omega$, the standard objective of off-policy learning is to maximize the expected action value function over the stationary distribution $\rho^{\beta}$, such that
\begin{equation}
    J_{\omega}(\theta) = \int_{s \in S} \rho^{\beta}(s\mid \omega) Q^{\mu_\theta}(s, \mu_{\theta}(s), \omega) \mathrm{d}s \enspace.\label{eq:jomega}
\end{equation}
Here, $\rho^{\beta}(s \mid \omega) = \lim_{T\to\infty}\frac{1}{T} \sum_{t=0}^{T-1} \int_{s_0} q_{\omega, \beta}^{t}(s \mid s_0) p_\omega^0(s_0) \mathrm{d}s_0$ denotes the stationary distribution under $\beta$ and a fixed $\omega \in \Omega$,
where the step-$t$ transition probability density $q_{\omega,\beta}^{t}$ is defined as
$q_{\omega,\beta}^{1}(s' \mid s) =\textstyle \int_{a \in A} q_{\omega}(s' \mid s, a) \beta(a \mid s) \mathrm{d}a$ and $q_{\omega,\beta}^{t}(s' \mid s) =\textstyle \int_{\bar{s} \in S} q_{\omega,\beta}^{t-1}(\bar{s} \mid s) q_{\omega,\beta}^{1}(s' \mid \bar{s}) \mathrm{d}\bar{s}$.

We assume that the agent can interact with $\mathcal{M}_{\omega}$ for any $\omega \in \Omega$ during the training and can change $\omega$ after every episode, that is, when $h_t = 1$.
Our objective is to obtain the $\mu_\theta$ that maximizes $J_\omega(\theta)$ under the worst environment $\omega \in \Omega$.
Hence, we tackle the max-min optimization problem requiring $\max_{\theta \in \Theta} \min_{\omega \in \Omega} J_{\omega}(\theta)$.

\section{\proposed} \label{sec:proposed}
We propose an off-policy actor-critic approach to obtain a policy that maximizes the worst-case performance on $\mathcal{M}_\Omega$, called \proposed.
Our approach is based on TD3, an off-policy actor-critic algorithm for a deterministic target policy.
In TD3, the critic network $Q_\phi$ parameterized by $\phi \in \Phi$ is trained to approximate the $Q^{\mu_\theta}$ of $\mu_\theta$, whereas the actor network models $\mu_\theta$ with parameter $\theta \in \Theta$ and is trained to maximize $J_\omega(\theta)$.
Moreover, $\omega$ is not considered and is fixed during the training.
In this study, to obtain the $\mu_\theta$ that maximizes the performance under the worst $\omega$, we extend TD3 by formulating the objective as a maximin problem and introducing a simultaneous gradient ascent descent approach.

The main difficulty in the worst-case performance maximization is in estimating the worst-case performance. Because the objective is expected to be non-concave with respect to (w.r.t.) the uncertainty parameter because of deep actor-critic networks, solving this problem is considered intractable in general~\cite{DBLP:conf/stoc/DaskalakisSZ21}.
To stabilize the worst-case performance estimation, we introduce various techniques: multiple uncertainty parameters, an uncertainty parameter refresh strategy, and an uncertainty parameter sampler.

\paragraph{Formulation}
We can formulate our objective as the following tri-level optimization:
\begin{align}
    \max_{\theta \in \Theta} \min_{\hat{\omega} \in \Omega} J(\theta, \hat{\omega}; \phi^*) \quad \text{s.t.}\quad \phi^* \in \argmin_{\phi \in \Phi} L(\phi; \theta) \enspace,\label{eq:tri}
\end{align}
where $J$ and $L$ are defined below. Note that this tri-level optimization formulation is an extension of the bi-level optimization formulation of the actor-critic approach proposed in \cite{DBLP:journals/corr/PfauV16}.

We introduce a probability density $\alpha: \Omega \to \R$, from which an $\omega$ to be used in the interaction during the training phase is drawn for each episode.
The relationship between $\hat{\omega}$ and $\alpha$ is similar to that between $\theta$ and $\beta$: namely, $\hat{\omega}$ and $\theta$ are the parameters to be optimized while $\alpha$ and $\beta$ are introduced for exploration.
Let $\rho_{\alpha}^{\beta}(s) = \int_{\omega \in \Omega} \rho^{\beta}(s\mid \omega) \alpha(\omega) \mathrm{d}\omega$ and $\rho_{\alpha}^{\beta}(s, a, \omega) = \beta(a \mid s) \rho^{\beta}(s\mid \omega) \alpha(\omega)$ be the stationary distribution of $s$ and the joint stationary distribution of $(s, a, \omega)$, respectively, under $\alpha$ and $\beta$.

The critic loss function $L(\phi; \theta)$ is designed to simulate the Q-learning algorithm.
Let $T_{\mu_\theta}$ be the function satisfying $T_{\mu_\theta}[Q](s, a, \omega) = r_\omega(s, a) + \gamma \int_{s'\in S} Q(s', \mu_\theta(s'), \omega) q_\omega(s'\mid s, a) \mathrm{d}s'$.
Then, \eqref{eq:q-recursive} states that $Q^{\mu_\theta}$ is the solution to $Q = T_{\mu_\theta}[Q]$.
Therefore, the critic is trained to minimize the difference between $Q_\phi$ and $T_{\mu_\theta}[Q_\phi]$.
This is achieved by minimizing
\begin{equation}
L(\phi; \theta) := \int_{s \in S}\int_{a \in A}\int_{\omega \in \Omega} (T_{\mu_\theta}[Q_{\phi}](s, a, \omega) - Q_{\phi}(s, a, \omega))^2 \rho_{\alpha}^{\beta}(s, a, \omega) \mathrm{d}s\mathrm{d}a\mathrm{d}\omega \enspace.\label{eq:l}
\end{equation}

The max-min objective function of the actor network
\begin{align}
    J(\theta, \hat{\omega}; \phi) &:= \int_{s \in S} Q_{\phi}(s, \mu_{\theta}(s), \hat{\omega}) \rho_{\alpha}^{\beta}(s) \mathrm{d}s \label{eq:j}
\end{align}
measures the performance of $\mu_\theta$ under the uncertainty parameter $\hat{\omega} \in \Omega$ approximated using the critic network $Q_{\phi}$ instead of the ground truth $Q^{\mu_\theta}$.
Similar to the standard actor-critic approach, we expect $Q_\phi$ to approach $Q^{\mu_\theta}$ as the critic loss is minimized.
Therefore, we expect $J(\theta, \hat{\omega}; \phi^*)$ approximates $\int_{s \in S} Q^{\mu_\theta}(s, \mu_{\theta}(s), \hat{\omega}) \rho_{\alpha}^{\beta}(s) \mathrm{d}s$ once we obtain $\phi^* \approx \argmin_{\phi \in \Phi} L(\phi; \theta)$.

Particularly, even if $Q_{\phi^*} = Q^{\mu_\theta}$, our objective function $J(\theta, \hat{\omega}; \phi^*)$ differs from $J_{\omega}(\theta)$ with $\omega = \hat{\omega}$ in \eqref{eq:jomega}, in that the stationary distribution $\rho^\beta(s \mid \omega)$ under fixed $\omega$ in $J_{\omega}$ is replaced with $\rho_{\alpha}^\beta(s)$ under $\omega \sim \alpha$.
However, this change allows us to effectively utilize the replay buffer, which stores the interaction history, to approximate the objective $J(\theta, \hat{\omega}; \phi)$.
Moreover, if $\alpha$ is concentrated at $\hat{\omega}$, $J(\theta, \hat{\omega}; \phi^*)$ coincides with $J_{\omega = \hat{\omega}}(\theta)$.

\paragraph{Algorithmic Framework}
The framework of \proposed{} is designed to solve the tri-level optimization \eqref{eq:tri}.
The overall framework of \proposed{} follows that of TD3.
In each episode, an uncertainty parameter $\omega$ is sampled from $\alpha$.
The training agent interacts with the environment $\mathcal{M}_\omega$ using behavior policy $\beta$.
At each time step $t$, the transition $(s_t, a_t, r_t, s_t', h_t, \omega)$ is stored in the replay buffer, which is denoted by $B$.
The critic network $Q_\phi$ is trained at every time step, to minimize \eqref{eq:l}, with a mini-batch being drawn uniformly randomly from $B$.
The actor network $\mu_\theta$ and the worst-case uncertainty parameter $\hat{\omega}$ are trained in every $T_\text{freq}$ step to optimize \eqref{eq:j}.
Note that a uniform random sample $(s, a, \omega)$ taken from $B$ can be regarded as a sample from the stationary distribution $\rho_\alpha^\beta(s, a, \omega)$.
Similarly, $s$ taken from $B$ is regarded as a sample from $\rho_\alpha^\beta(s)$.
These facts allow approximation of the expectations in \eqref{eq:l} and \eqref{eq:j} using the Monte Carlo method with mini-batch samples uniformly randomly taken from $B$.
An algorithmic overview of the proposed method is summarized in \Cref{alg:proposed_minimum_configuration}.
We have underlined the differences from the general off-policy actor-critic method in the episodic settings.
A detailed description of \proposed{} is summarized in \Cref{alg:proposed} in \Cref{apdx:algo}.
As shown from \Cref{alg:proposed_minimum_configuration}, the differences between the proposed method and the general off-policy actor-critic are as follows: (1) the introduction of an uncertainty parameter sampler, (2) definition and updating method of critic, and (3) updating method of actor and uncertainty parameters.

\begin{algorithm}[h]
  \caption{Algorithmic overview of the proposed method}
  \label{alg:proposed_minimum_configuration}
  \begin{algorithmic}[1]
  \renewcommand{\algorithmicrequire}{\textbf{Input:}}
  \renewcommand{\algorithmicensure}{\textbf{Output:}}
  \STATE \underline{Draw uncertainty parameter $\omega \sim \alpha_0$}
  \STATE Observe initial state $s \sim p_\omega^0$
  \FOR {$\mathrm{t}=1$ to $T_\text{max}$}
    \STATE \texttt{\# interaction}
    \STATE Select action $a \sim \beta_t(s)$
    \STATE Interact with $\mathcal{M}_\omega$ with $a$, observe next state $s'$, immediate reward $r$ and termination flag $h$
    \STATE Store transition tuple $(s, a, r, s', h, \omega)$ in $B$
    \IF {$h = 1$}
        \STATE \underline{Reset uncertainty parameter $\omega \sim \alpha_t$}
        \STATE Observe initial state $s \sim p_\omega^0$
    \ELSE
        \STATE Update current state $s \leftarrow s'$
    \ENDIF
    \STATE \texttt{\# learning}
    \STATE Sample mini-batch $\{(s_i, a_i, r_i, s_i', h_i, \omega_i)\}$ of $\mathrm{M}$ transitions uniform-randomly from $B$
    \STATE \underline{Update the critic network by optimizing \Cref{eq:tildel}}
    \STATE \underline{Update the actor network and the uncertainty parameter by optimizing \Cref{eq:tildej}}
    \STATE \underline{Update uncertainty parameter sampler $\alpha$}
  \ENDFOR
  \end{algorithmic}
\end{algorithm}

\paragraph{Critic Update}
The critic network update, namely, minimization of \eqref{eq:l}, is performed  by implementing the TD error-based approach.
The concept is as follows.
Let $\{(s_i, a_i, r_i, s_i', \omega_i)\}_{i=1}^{M} \subset B$ be the mini-batch, and let $\theta_t$ and $\phi_t$ be the actor and critic parameters at time step $t$, respectively.
For each tuple, $T_{\mu_{\theta_{t}}}[Q_{\phi_{t}}](s_i, a_i, \omega_i)$ is approximated by $y_i = r_i + \gamma \cdot Q_{\phi_t}(s_i', \mu_{\theta_t}(s_i'), \omega_i)$.
Then, $L(\phi_t; \theta_t)$ in \eqref{eq:l} is approximated by the mean square error $\tilde{L}(\phi_t)$, such that
\begin{equation}
    \tilde{L}(\phi_t) = \frac{1}{M} \sum_{i=1}^{M} (y_i - Q_{\phi_t}(s_i, a_i, \omega_i))^2 \enspace.\label{eq:tildel}
\end{equation}
Then, the update follows $\phi_{t+1} = \phi_t - \lambda_\phi \nabla_\phi \tilde{L}(\phi_t) $.

We incorporate the techniques introduced in TD3 to stabilize the training: clipped double Q-learning, target policy smoothing regularization, and target networks.
Additionally, we introduce the smoothing regularization for the uncertainty parameter so that the critic is smooth w.r.t.~$\hat{\omega}$.
\Cref{alg:critic} in \Cref{apdx:algo} summarizes the critic update and the details of each technique are available in \cite{fujimoto2018addressing}.

\paragraph{Actor Update}
Updating of the actor as well as the uncertainty parameter, namely, maximin optimization of \eqref{eq:j}, is performed via simultaneous gradient ascent descent \cite{Nagarajan2017}, summarized in \Cref{alg:actor} in \Cref{apdx:algo}.
Let $\phi_t$, $\theta_t$ and $\hat{\omega}_t$ be the critic, policy, and the uncertainty parameters, respectively, at time step $t$.
Instead of the optimal $\phi^*$ in \eqref{eq:tri}, we use $\phi_t$ as its estimate.
With the mini-batch $\{s_i\}_{i=1}^{M} \subset B$, we approximate the max-min objective \eqref{eq:j} follows:
\begin{align}
    \tilde{J}_t(\theta, \hat{\omega}) &= \frac{1}{M}\sum_{i=1}^{M} Q_{\phi_t}(s_i, \mu_{\theta}(s), \hat{\omega}) \enspace. \label{eq:tildej}
\end{align}
We update $\theta$ and $\hat{\omega}$ as $\theta_{t+1} = \theta_{t} + \lambda_\theta \nabla_\theta \tilde{J}_t(\theta_t, \hat{\omega}_t)$ and $\hat{\omega}_{t+1} = \hat{\omega}_{t} - \lambda_{\omega} \nabla_{\hat{\omega}} \tilde{J}_t(\theta_t, \hat{\omega}_t)$, respectively.

For higher-stability performance, we introduce multiple uncertainty parameters.
The motivation is twofold.
One is to deal with multiple local minima of $J_t(\theta, \hat{\omega}; \phi^*)$ w.r.t.~$\hat{\omega}$.
As the critic network becomes generally non-convex, there may exist multiple local minima of $J_t(\theta, \hat{\omega}; \phi^*)$ w.r.t.~$\hat{\omega}$.
Once $\hat{\omega}$ is stacked at a local minimum point, e.g., $\hat{\omega}^*$, $\theta$ may be trained to be robust around a neighborhood of $\hat{\omega}^*$ and to perform poorly outside that neighborhood.
The other is that the maximin solution $(\theta^*, \phi^*)$ of $J_t(\theta, \hat{\omega}; \phi^*)$ is not a saddle point, which occurs when the objective is non-concave in $\hat{\omega}$~\cite{DBLP:conf/icml/JinNJ20}.
Here, more than one $\hat{\omega}$ is necessary to approximate the worst-case performance $\min_{\hat{\omega}} J_t(\theta, \hat{\omega}; \phi^*)$ around $(\theta^*, \phi^*)$ and a standard simultaneous gradient ascent descent method fails to converge.
To relax this defect of simultaneous gradient ascent descent methods, we maintain multiple candidates for the worst $\hat{\omega}$, denoted by $\hat{\omega}_1,\dots,\hat{\omega}_N$.
Therefore, replacing $\min_{\hat{\omega}} J(\theta, \hat{\omega}; \phi^*)$ with $ \min_{k=1,\dots,N} J(\theta, \hat{\omega}_k; \phi^*)$ in \eqref{eq:j} has no effect on the optimal $\theta$; however, this change does affect the training behavior.
Our update follows the simultaneous gradient ascent descent on $\min_{k=1,\dots,N} J_t(\theta, \hat{\omega}_k)$: $\theta \leftarrow \theta + \lambda_\theta \nabla_\theta (\min_{k=1,\dots,N} J_t(\theta, \hat{\omega}_k))$ and $\hat{\omega}_k \leftarrow \hat{\omega}_k - \lambda_\omega \nabla_{\hat{\omega}_k}( \min_{\ell=1,\dots,N} J_t(\theta, \hat{\omega}_\ell))$ for $k = 1, \dots, N$.
Hence, $\theta$ is updated against the worst $\hat{\omega}_k$, and only the worst $\hat{\omega}_k$ is updated because the gradient w.r.t. the other $\hat{\omega}_k$ is zero.

For multiple uncertainty parameters to be effective, they must be distant from each other.
Moreover, all are expected to be selected as the worst parameters with non-negligible frequencies.
Otherwise, the advantage of having multiple uncertainty parameters is lessened.
From this perspective, we introduce a refreshing strategy for uncertainty parameters.
Namely, we resample $\hat{\omega}_k \sim \mathcal{U}(\Omega)$ if one of the following scenarios is observed: there exists $\hat{\omega}_{\ell}$ such that the distance $d_{\hat{\omega}}(\hat{\omega}_k, \hat{\omega}_{\ell}) \leq d_\mathrm{thre}$;  the frequency $p_k$ of $\hat{\omega}_k$ being selected as the worst-case during the actor update is no greater than $p_\mathrm{thre}$.

\paragraph{Uncertainty Parameter Sampler}
The uncertainty parameter sampler $\alpha$ controls the exploration-exploitation trade-off in the uncertainty parameter.
Exploration in $\Omega$ is necessary to train the critic network. If the critic network is not well trained over $\Omega$, it is difficult to locate the worst $\hat{\omega}$ correctly.
On the other hand, for $J(\theta, \hat{\omega}; \phi^*)$ in \eqref{eq:j} to coincide with $J_\omega(\theta)$ for $\omega = \hat{\omega}$ in \eqref{eq:jomega}, we require $\alpha$ to be concentrated at $\hat{\omega}$.
Otherwise, the optimal $\theta$ for $J(\theta, \hat{\omega}; \phi^*)$ may deviate from that of $J_\omega(\theta)$.
We design $\alpha$ as follows.
For the first $T_\text{rand}$ steps, we sample the uncertainty parameter $\omega$ uniformly randomly on $\Omega$, i.e., $\alpha = \mathcal{U}(\Omega)$.
Then, we set $\alpha = \sum_{k=1}^{N} p_k \cdot \mathcal{N}(\hat{\omega}_k, \Sigma_\omega)$, where $\Sigma_\omega$ is the predefined covariance matrix.
We decrease $\Sigma_\omega$ as the time step increases.
The rationale behind the gradual decrease of $\Sigma_\omega$ is that the training of the critic network is still important after $T_\text{rand}$ to make the estimation of the worst-case uncertainty parameter accurate.
Details are provided in \Cref{apdx:ex}.

\section{Related Work}\label{sec:related}

Methods to handle model misspecification include (1) robust policy searching~\cite{morimoto2005robust,DBLP:conf/icml/PintoDSG17,DBLP:conf/aaai/MankowitzMBPM18,li2019robust,iyengar2005robust,Mankowitz2020Robust}, (2) transfer learning~\cite{DBLP:journals/jmlr/TaylorS09,DBLP:conf/iclr/YuLT19,DBLP:conf/iclr/RajeswaranGRL17}, (3) domain randomization (DR)~\cite{tobin2017domain}, and (4) approaches minimizing worst-case sub-optimality gap~\cite{DBLP:conf/nips/0001JVBRCL20,DBLP:conf/nips/LinTYM20}.
Robust policy searching includes methods that aim to obtain the optimal policy under the worst possible disturbance or model misspecification, i.e., maximization of $\min_{\omega \in \Omega} J_\omega(\theta)$ for an explicitly or implicitly defined $\Omega$.
Transfer learning is a method in which a policy is trained on source tasks and then fine-tuned through interactions while performing the target task.
DR is a method in which a policy is trained on source tasks that are sampled randomly from $\mathcal{M}_\Omega$.
There are two types of DR approaches: vanilla and guided. In vanilla DR~\cite{tobin2017domain}, the policy is trained on source tasks that are randomly sampled from a pre-defined distribution on $\Omega$.
In guided DR~\cite{yu2018policy}, the policy is trained on source tasks; however, the $\omega$ distribution is guided towards the target task.
Because we do not assume access to the target task, transfer learning and many guided DR approaches are outside the scope of this work.
Some guided DR approaches, such as active domain randomization \cite{Mehta2019ActiveDR}, do not access the target task or consider worst-case optimization either.
Vanilla~DR can be applied to the setting considered here.
However, the objective of vanilla DR differs from the present aim, i.e., to maximize the performance averaged over $\Omega$, namely, $\E_{\omega}[J_\omega(\theta) \mid \omega \sim \mathcal{U}(\Omega)]$, if the sampling distribution is $\mathcal{U}(\Omega)$.
The approaches minimizing the worst-case sub-optimality gap \cite{DBLP:conf/nips/0001JVBRCL20,DBLP:conf/nips/LinTYM20} do not optimize the worst-case performance, instead attempt to obtain a policy that generalizes well on the uncertainty parameter set while avoiding too conservative performance, which often attributes to the worst-case performance optimization.

Some robust policy search methods, some adopt an adversarial approach to policy optimization.
For example, Robust Adversarial Reinforcement Learning (RARL) \cite{DBLP:conf/icml/PintoDSG17} models the disturbance caused by an external force produced by an adversarial agent, and alternatively trains the protagonist and adversarial agents.
Robust Reinforcement Learning (RRL)~\cite{morimoto2005robust} similarly models the disturbance but has not been applied to the DRL framework.
Minimax Multi-Agent Deep Deterministic Policy Gradient (M3DDPG)~\cite{li2019robust} has been designed to obtain a robust policy for multi-agent settings.
This method is applicable to the setting targeted in this study if only two agents are considered.
The above approaches frame the problem as a zero-sum game between a protagonist agent attempting to optimize $\mu_\theta$ and an adversarial agent attempting to minimize the protagonist's performance, hindering the protagonist by generating the worst possible disturbance.
Adv-Soft Actor Critic (Adv-SAC)~\cite{DBLP:conf/icra/JianYGL021} learns policies that are robust to both internal disturbances in the robot's joint space and those from other robots.
Recurrent SAC (RSAC)~\cite{DBLP:journals/corr/abs-2103-15370} introduces POMDPs to treat the uncertainty parameter as an unobservable state.
A DDPG based approach robustified by applying stochastic gradient langevin dynamics is proposed under the noisy robust MDP setting \cite{DBLP:conf/nips/KamalarubanHHRS20}.
However, it has been reported in \cite{DBLP:journals/corr/abs-2103-15370,DBLP:conf/nips/KamalarubanHHRS20} that their worst-case performances are sometimes even worse than their baseline non-robust approaches.
State-Adversarial-DRL (SA-DRL)~\cite{DBLP:conf/nips/0001CX0LBH20}, and alternating training with learned adversaries (ATLA)~\cite{DBLP:conf/iclr/ZhangCBH21} improve the robustness of DRL agents by using an adversary that perturbs the observations in  SAMDP framework that characterizes the decision-making problem in an adversarial attack on state observations.
They do not address the model misspecification in the reward and transition functions.

Other approaches attempt to estimate the robust value function, i.e., a value function under the worst uncertainty parameter.
Among them, Robust Dynamic Programming (DP)~\cite{iyengar2005robust} is a DP approach, while the Robust Options Policy Iteration (ROPI)~\cite{DBLP:conf/aaai/MankowitzMBPM18} incorporates robustness into option learning \cite{SUTTON1999181}, which allows agents to learn both hierarchical policies and their corresponding option sets.
ROPI is a type of Robust MDP approach~\cite{DBLP:journals/mor/WiesemannKR13,DBLP:conf/nips/LimXM13}.
Robust Maximum A-posteriori Policy Optimization (R-MPO)~\cite{Mankowitz2020Robust} incorporates robustness in MPO~\cite{abdolmaleki2018maximum}.

Neither Robust MDP nor R-MPO require interaction with $M_\omega$ for an arbitrary $\omega \in \Omega$.
This can be advantageous in scenarios where the design of a simulator valid over $\Omega$ is tedious.
However, these methods typically require additional assumptions for computating the worst-case for each value function update, such as the finiteness of the state-action space and/or the finiteness of $\Omega$.
To apply R-MPO to our setting, finite scenarios from $\Omega$ for training are sampled, but the choice of the training scenarios affects the worst-case performance.

Additionally, to optimize the worst-case performance in the field of offline RL~\cite{DBLP:conf/nips/XieCJMA21,DBLP:conf/nips/ZanetteWB21,DBLP:conf/icml/XieJ21}, offline RL attempts to obtain robust measures by introducing the principle of pessimism.
Some studies that introduce the principle of pessimism in the model-free context~\cite{DBLP:conf/nips/XieCJMA21,DBLP:conf/icml/XieJ21} and in the actor-critic context~\cite{DBLP:conf/nips/ZanetteWB21}.
However, these do not compete with this study due to differences in motivation and because offline RL requires additional assumptions in the datasets~\cite{DBLP:conf/nips/ZanetteWB21,DBLP:conf/icml/XieJ21} and realizability~\cite{DBLP:conf/nips/XieCJMA21}.

\Cref{table:robust_rrl} in \Cref{apdx:related} summarizes the robust policy search method taxonomy.
Our proposed approach, \proposed{}, considers the uncertainties in both the reward function and transition probability.
The state and action spaces, as well as the uncertainty set, are assumed to be continuous.
\Cref{table:def_q} in \Cref{apdx:related} compares the related methods considering their action value function definitions.
Methods that do not explicitly utilize the action value function are interpreted from the objective functions.
Both M3DDPG and \proposed{} maintain the value function for the tuple of $(s, a, \omega)$.
However, M3DDPG takes the worst $\omega'$ on the right-hand side, yielding a more conservative policy than \proposed{} for our setting because we assume a stationary uncertainty model, whereas M3DDPG assumes a time-varying uncertainty model in nature \cite{DBLP:journals/ior/NilimG05}.
The value functions in RARL and \proposed{} are identical if $\omega$ is fixed.
Particularly, $\omega$ is not introduced to the RARL value function because it does not simultaneously optimize $\mu_\theta$ and the estimate of the worst $\omega$.
RARL repeats the $\theta$ and $\hat{\omega}$ optimizations alternatively.
In contrast, \proposed{} optimizes $\theta$ and $\hat{\omega}$ simultaneously in a manner similar to the training processes of generative adversarial networks (GANs)~\cite{NIPS2014_5ca3e9b1}.
Hence, we require an action value for each $\omega \in \Omega$.

A difference exists between the optimization strategies of RARL and \proposed{}.
As noted in the aforementioned section, both methods attempt to maximize the worst-case performance $\min_{\omega \in \Omega} J_\omega(\theta)$.
Conceptually, RARL repeats $\theta \leftarrow \argmax_{\theta} J_\omega(\theta)$ and $\omega \leftarrow \argmin_{\omega} J_\omega(\theta)$.
However, this optimization strategy fails to converge even if the objective function $(\theta, \omega) \mapsto J_{\omega}(\theta)$ is concave-convex.
As an example, consider the function $(x, y) \mapsto y^2 - x^2 + \alpha xy$.
The RARL optimization strategy reads $x \leftarrow (\alpha/2) y$ and $y \leftarrow - (\alpha/2) x$, which causes divergence if $\alpha > 2$\footnote{A simple way to mitigate this issue would be to early-stop the optimization for each step of RARL. However, the problem remains.
The protagonist agent in RARL does not consider the uncertainty parameter in the critic.
This leads to a non-stationary training environment for both protagonist and adversarial agents.
As the learning in a non-stationary environment is generally difficult~\cite{pmlr-v97-iqbal19a}, RARL will be unstable.
The instability of RARL has also been investigated in the linear-quadratic system settings~\cite{DBLP:conf/nips/ZhangHB20}.
}.
Alternating updates are also employed in other approaches such as Adv-SAC, SA-DRL, and ATLA.
These share the same potential issue as RARL.
\proposed{} attempts to alleviate this divergence problem by applying the gradient-based max-min optimization method, which has been employed in GANs and other applications and analyzed for its convergence~\cite{DBLP:conf/nips/MeschederNG17,LiangS19}.

\section{Experiments} \label{sec:ex}
In this study, we conducted experiments on the optimal control problem using MuJoCo environments.
Hence, we demonstrated the problems of existing methods and assessed the worst-case performance and average performance of the policy trained by \proposed{} for different continuous control tasks.

\paragraph{Baseline Methods}
We summarize the baseline methods adapted to our experiment setting, namely, DR, RARL, and M3DDPG.

\underline{DR}:
The objective of DR is to maximize the expected cumulative reward for the distribution $\alpha = \mathcal{U}(\Omega)$ of $\omega$.
In each training episode, $\omega$ is drawn randomly from $\mathcal{U}(\Omega)$, and the agent neglects $\omega$ when training and performing the standard DRL. For a fair comparison, we implemented DR with TD3 as the baseline DRL method.

\underline{RARL}:
We adapted RARL to our scenario by setting $\mu_{\hat{\omega}}: s \mapsto \hat{\omega}$ to the antagonist policy.
RARL was regarded as optimizing \eqref{eq:jomega} for the worst $\omega \in \Omega$; hence, the objective was the same as that of \proposed{}.
Particularly, the main technical difference between \proposed{} and RARL is in the optimization strategy as described in \Cref{sec:related}.
The original RARL is implemented with Trust Region Policy Optimization (TRPO)~\cite{schulman2015trust}, but for a fair comparison, we implemented it with DDPG, denoted as RARL (DDPG).
The experimental results of RARL (TRPO), the original RARL, are provided in \Cref{apdx:RARL_TRPO}.

\underline{M3DDPG}:
By considering a two-agent scenario and a state-independent policy $\mu_{\hat{\omega}}: s \mapsto \hat{\omega}$ as the opponent-agent's policy, we adapted M3DDPG to our setting.
M3DDPG is different from M2TD3 (and M2-DDPG below) even under a state-independent policy as described in \Cref{sec:related}.
Because of the difficulty in replacing DDPG with TD3 in the M3DDPG framework, we used DDPG as the baseline DRL.

Additionally, for comparison, we implemented our approach with DDPG instead of TD3, denoted by M2-DDPG.
We also implemented a variant of \proposed{}, denoted as SoftM2TD3, performing a ``soft'' worst-case optimization to achieve better average performance while considering the worst-case performance.
The detail of SoftM2TD3 is described in \Cref{apdx:softm2td3}.

\begin{table*}[!t]
\begin{center}
\caption{Avg.\ $\pm$ std.\ error of worst-case performance $R_\text{worst}(\mu)$ over 10 trials for each approach}
\label{table:results_mujoco}
\scalebox{0.7}{
\begin{tabular}{|c|c|c|c|c|c|c|}
\bottomrule
Environment                & \proposed{} &  SoftM2TD3 & M2-DDPG & M3DDPG & RARL (DDPG) & DR (TD3)            \\ 
\toprule\bottomrule
Ant 1 $(\times 10^{3})$& $ 3.84 \pm 0.10$ & $ \bm{4.08 \pm 0.15} $ & $ 1.28 \pm 0.19$ & $ 0.49 \pm 0.12$ & $ -1.24 \pm 0.10$ & $3.51 \pm 0.08$\\ \hline
Ant 2 $(\times 10^{3})$& $ \bm{4.13 \pm 0.11}$ & $ 3.92 \pm 0.14$ & $ 0.95 \pm 0.20$ & $ -0.25 \pm 0.13$ & $ -1.77 \pm 0.09$ & $ 1.64 \pm 0.13 $\\ \hline
Ant 3 $(\times 10^{3})$& $ \bm{0.10 \pm 0.10}$ & $ 0.07 \pm 0.20$ & $ -1.13 \pm 0.28$ & $ -1.38 \pm 0.22$ & $ -2.38 \pm 0.07$ & $ -0.32 \pm 0.03$\\
\toprule\bottomrule
HalfCheetah 1 $(\times 10^{3})$& $ 3.14 \pm 0.10$ & $ \bm{3.24 \pm 0.08}$ & $ 2.24 \pm 0.25$ & $ -0.13 \pm 0.12$ & $ -0.55 \pm 0.02$ & $ 3.19 \pm 0.08$\\ \hline
HalfCheetah 2 $(\times 10^3)$ & $ 2.61 \pm 0.16 $ & $ \bm{2.82 \pm 0.16} $ & $ 2.54 \pm 0.23 $ & $ -0.58 \pm 0.06 $ & $ -0.70 \pm 0.05 $ & $ 2.12 \pm 0.13 $\\ \hline
HalfCheetah 3 $(\times 10^3)$ & $ 0.93 \pm 0.21 $ & $ \bm{1.53 \pm 0.23} $ & $ 1.20 \pm 0.22 $ & $ -0.66 \pm 0.08 $ & $ -0.81 \pm 0.07 $ & $ 1.09 \pm 0.06 $ \\
\toprule\bottomrule
Hopper 1 $(\times 10^2)$ & $ \bm{6.21 \pm 0.45} $ & $ 5.98 \pm 0.23 $ & $ 5.38 \pm 0.43 $ & $ 4.14 \pm 0.60 $ & $ 3.32 \pm 0.78 $ & $ 5.28 \pm 2.55 $ \\ \hline
Hopper 2 $(\times 10^2)$ & $ 5.33 \pm 0.28 $ & $ \bm{5.79 \pm 0.29} $ & $ 4.30 \pm 0.57 $ & $ 2.58 \pm 0.29 $ & $ 3.34 \pm 0.89 $ & $ 4.68 \pm 0.15 $ \\ \hline
Hopper 3 $(\times 10^2)$ & $ \bm{2.84 \pm 0.25} $ & $ 1.98 \pm 0.22 $ & $ 2.25 \pm 0.29 $ & $ 0.73 \pm 0.11 $ & $ 1.64 \pm 0.46 $ & $ 2.10 \pm 0.35 $ \\
\toprule\bottomrule
HumanoidStandup 1 $(\times 10^4)$ & $ 9.33 \pm 0.70 $ & $ 9.49 \pm 0.81 $ & $ 8.09 \pm 0.92 $ & $ 8.00 \pm 0.78 $ & $ 5.29 \pm 0.45 $ & $ \bm{9.68 \pm 0.60} $ \\ \hline
HumanoidStandup 2 $(\times 10^4)$ & $ 6.50 \pm 0.70 $ & $ \bm{7.94 \pm 0.90} $ & $ 6.24 \pm 0.54 $ & $ 6.37 \pm 0.72 $ & $ 5.78 \pm 0.73 $ & $ 7.31 \pm 0.78 $ \\ \hline
HumanoidStandup 3 $(\times 10^4)$ & $ \bm{6.20 \pm 0.64} $ & $ 5.99 \pm 0.37 $ & $ 5.96 \pm 0.58 $ & $ 6.01 \pm 0.38 $ & $ 5.54 \pm 0.76 $ & $ 5.41 \pm 0.34 $ \\
\toprule\bottomrule
InvertedPendulum 1 $(\times 10^2)$ & $ \bm{8.22 \pm 1.13} $ & $ 6.53 \pm 1.36 $ & $ 6.49 \pm 1.33 $ & $ 1.09 \pm 0.71 $ & $ 1.53 \pm 0.64 $ & $ 3.18 \pm 1.10 $ \\ \hline
InvertedPendulum 2 $(\times 10^2)$ & $ \bm{3.56 \pm 1.32} $ & $ 1.36 \pm 0.30 $ & $ 1.10 \pm 0.62 $ & $ 0.02 \pm 0.00 $ & $ 0.02 \pm 0.00 $ & $ 0.57 \pm 0.02 $ \\
\toprule\bottomrule
Walker 1 $(\times 10^3)$ & $ 2.83 \pm 0.39 $ & $ \bm{3.02 \pm 0.22} $ & $ 1.19 \pm 0.17 $ & $ 0.89 \pm 0.18 $ & $ 0.09 \pm 0.02 $ & $ 2.19 \pm 0.40 $ \\ \hline
Walker 2 $(\times 10^3)$ & $ \bm{3.14 \pm 0.39} $ & $ 2.64 \pm 0.43 $ & $ 0.85 \pm 0.12 $ & $ 0.39 \pm 0.11 $ & $ 0.06 \pm 0.04 $ & $ 2.31 \pm 0.50 $ \\ \hline
Walker 3 $(\times 10^3)$ & $ 1.94 \pm 0.40 $ & $ \bm{2.00 \pm 0.35} $ & $ 0.82 \pm 0.13 $ & $ 0.28 \pm 0.09 $ & $ 0.00 \pm 0.02 $ & $ 1.32 \pm 0.34 $ \\
\toprule\bottomrule
Small HalfCheetah 1 $(\times 10^3)$ & $ 5.27 \pm 0.12 $ & $ 5.07 \pm 0.14 $ & $ 4.51 \pm 0.18 $ & $ 1.26 \pm 0.38 $ & $ -0.52 \pm 0.02 $ & $ \bm{6.76 \pm 0.18} $ \\ \hline
Small Hopper 1 $(\times 10^3)$ & $ 2.88 \pm 0.32 $ & $ 2.23 \pm 0.32 $ & $ 1.40 \pm 0.19 $ & $ 1.39 \pm 0.21 $ & $ 0.51 \pm 0.12 $ & $ \bm{3.42 \pm 0.11} $ \\
\toprule
\end{tabular}
}
\end{center}
\end{table*}

\paragraph{Experiment Setting}
We constructed 19 tasks based on six MuJoCo environments and with 1-3 uncertainty parameters, as summarized in \Cref{table:environment_settings} in \Cref{apdx:ex}.
Here, $\Omega$ was defined as an interval, which mostly included the default $\omega$ values.

To assess the worst-case performance of the given policy $\mu$ under $\omega \in \Omega$, we evaluated the cumulative reward 30 times for each uncertainty parameter value $\omega_1,\dots,\omega_K \in \Omega$.
Here, $R_{k}(\mu)$ was defined as the cumulative reward on $\omega_k$ averaged over 30 trials.
Then, $R_\text{worst}(\mu) = \min_{1\leq k\leq K} R_{k}(\mu)$ was measured as an estimate of the worst-case performance of $\mu$ on $\Omega$.
We also report the average performance $R_\text{average}(\mu) = \frac{1}{K}\sum_{k=1}^{K} R_{k}(\mu)$.
A total of $K$ uncertainty parameters $\omega_1,\dots,\omega_K$ for evaluation were drawn as follows: for 1D $\omega$, we chose $K=10$ equally spaced points on the 1D interval $\Omega$; for 2D $\omega$, we chose 10 equally spaced points in each dimension of $\Omega$, thereby obtaining $K = 100$ points; and for 3D $\omega$, we chose 10 equally spaced points in each dimension of $\Omega$, thereby obtaining $K=1000$ points.

For each approach, we trained the policy 10 times in each environment.
The training time steps $T_\text{max}$ were set to 2M, 4M, and 5M for the senarios with 1D, 2D, and 3D uncertainty parameters, respectively.
The final policies obtained were evaluated for their worst-case performances.
For further details of the experiment settings, please refer to \Cref{apdx:ex}.
\paragraph{Comparison to Baseline Methods}
\begin{figure}[t]
  \begin{center}
    \includegraphics[trim=0 0 0 0, clip, width=\hsize] {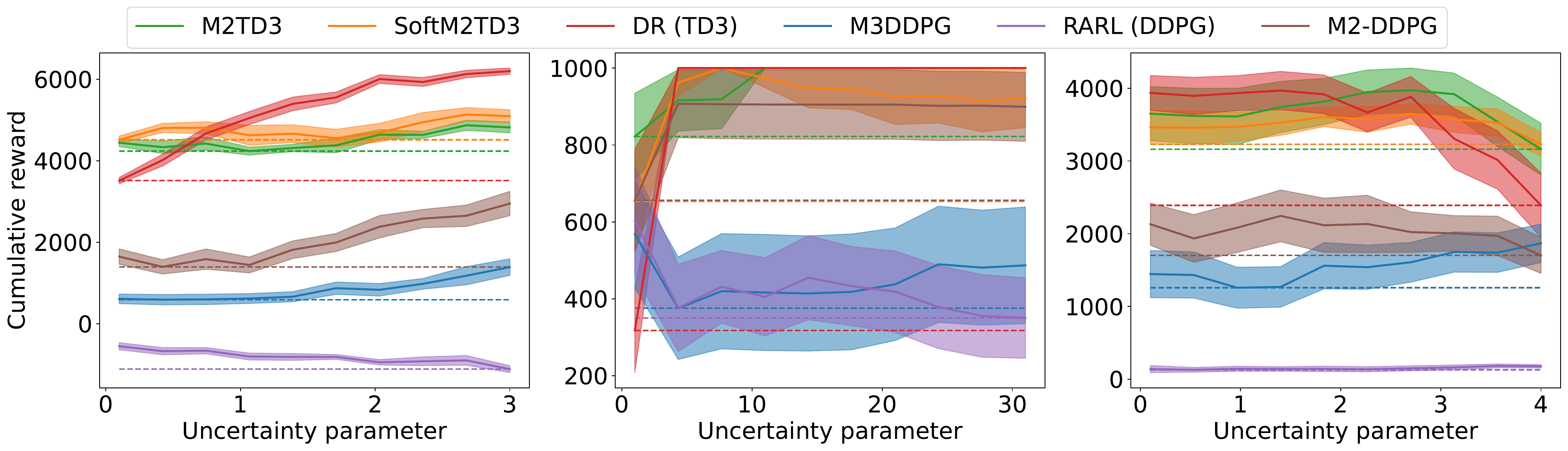}%
    \caption{Cumulative rewards under trained policies for each uncertainty parameter $\omega \in \Omega$. The average (solid line) and standard error (band) for each $\omega \in \Omega$, as well as the worst average value (dashed line) are shown.
    Left: Ant 1, 
    Middle: InvertedPendulum 1, 
    Right: Walker 1.
    }
  \label{fig:each_param}
 \end{center}
\end{figure}
\Cref{table:results_mujoco} summarizes the worst-case performances of the policies trained by \proposed{}, SoftM2TD3, M2-DDPG, M3DDPG, RARL, and DR.
See also \Cref{table:td3_reference} in \Cref{apdx:td3} for the results of TD3 trained on the reference uncertainty parameters as baselines and \Cref{fig:training_curve_worst_all} in \Cref{apdx:training_curve_all} for the learning curves.
In most cases, \proposed{} and SoftM2TD3 outperformed DR, and RARL, and M3DDPG.
\Cref{fig:each_param} shows the cumulative rewards of the policies trained by the six approaches for each of the $\omega$ values on the Ant 1, InvertedPendulum 1, and Walker 1 senarios. See also \Cref{fig:each_rewards_1_all} in \Cref{apdx:each_rewards_1_all} for other senarios.

\underline{Comparison with DR}:
DR does not attempt to optimize the worst-case performance.
In fact, it showed lower worst-case performance than \proposed{} and Soft\proposed{} in many scenarios because the obtained policy performs poorly on some uncertainty parameters while it performs well on average.
In the results of InvertedPendulum 1 shown in \Cref{fig:each_param}, for example, the policy obtained by DR exhibited a high performance for a wide range of $\omega \in \Omega$ but performed poorly for small $\omega$.
\proposed{} outperformed DR and the other baselines on those senarios.
However, in some scenarios, such as HumanoidStandup 1, Small HalfCheetah 1, and Small Hopper 1,
DR achieved a better worst-case performance than \proposed.
This outcome may be because the optimization of worst-case function values is generally more unstable than that of the expected function values.

\underline{Comparison with RARL}:
M2-DDPG outperformed RARL in most senarios, with this performance difference originating from the different optimization strategies.
As noted above, the optimization strategy employed in RARL often fails to converge.
Specifically, as shown in \Cref{fig:each_param}, RARL failed to optimize the policy not only in the worst-case but also on average in Ant 1 and Walker 1 senarios.
In the InvertedPendulum 1 senario, RARL could train the policy for some uncertainty parameters, but not for the worst-case uncertainty parameter.

\underline{Comparison with M3DDPG}:
In many cases, M2-DDPG outperformed M3DDPG.
Because M3DDPG considers a stronger adversary than necessary (the worst $\omega$ for each time step), it was too conservative for this experiment and exhibited lower performance.

\paragraph{Average Performance}
Although our objective is maximizing the worst-case performance, the average performance is also important in practice.
\Cref{table:results_mujoco_mean} compares the average performance of the six approaches.
Generally, DR achieved the highest average performance as expected.
However, interestingly, \proposed{} and SoftM2TD3 achieved competitive average performances to DR on several senarios such as Hopper 1, 2, and Walker 1, 2.
Moreover, \Cref{table:results_mujoco} compared with \Cref{table:results_mujoco_mean} shows that a few times higher average performance than worst-case performance on several senarios such as Ant 3 and Hopper 1--3.

\begin{table*}[!t]
\begin{center}
\caption{Avg.\ $\pm$ std.\ error of average performance $R_\text{average}(\mu)$ over 10 trials for each approach}
\label{table:results_mujoco_mean}
\scalebox{0.7}{
\begin{tabular}{|c|c|c|c|c|c|c|}
\bottomrule
Environment                & \proposed{} &  SoftM2TD3 & M2-DDPG & M3DDPG & RARL (DDPG) & DR (TD3)            \\ 
\toprule\bottomrule
Ant 1 $(\times 10^3)$ & $ 4.51 \pm 0.08 $ & $ 4.78 \pm 0.16 $ & $ 2.05 \pm 0.21 $ & $ 0.84 \pm 0.14 $ & $ -0.82 \pm 0.06 $ & $ \bm{5.25 \pm 0.10} $ \\ \hline
Ant 2 $(\times 10^3)$ & $ 5.44 \pm 0.05 $ & $ 5.56 \pm 0.01 $ & $ 3.03 \pm 0.19 $ & $ 1.86 \pm 0.38 $ & $ -1.00 \pm 0.06 $ & $ \bm{6.32 \pm 0.09} $ \\ \hline
Ant 3 $(\times 10^3)$ & $ 2.66 \pm 0.22 $ & $ 2.98 \pm 0.23 $ & $ 0.30 \pm 0.46 $ & $ -0.33 \pm 0.22 $ & $ -1.31 \pm 0.09 $ & $ \bm{3.62 \pm 0.11} $ \\
\toprule\bottomrule
HalfCheetah 1 $(\times 10^3)$ & $ 3.89 \pm 0.06 $ & $ 4.00 \pm 0.05 $ & $ 3.50 \pm 0.12 $ & $ 1.01 \pm 0.26 $ & $ -0.46 \pm 0.02 $ & $ \bm{5.93 \pm 0.18} $ \\ \hline
HalfCheetah 2 $(\times 10^3)$ & $ 4.35 \pm 0.05 $ & $ 4.52 \pm 0.07 $ & $ 3.91 \pm 0.08 $ & $ 0.77 \pm 0.12 $ & $ -0.08 \pm 0.05 $ & $ \bm{5.79 \pm 0.15} $ \\ \hline
HalfCheetah 3 $(\times 10^3)$ & $ 3.79 \pm 0.09 $ & $ 4.02 \pm 0.04 $ & $ 3.39 \pm 0.21 $ & $ 0.58 \pm 0.18 $ & $ -0.21 \pm 0.10 $ & $ \bm{5.54 \pm 0.16} $ \\
\toprule\bottomrule
Hopper 1 $(\times 10^3)$ & $ \bm{2.68 \pm 0.11} $ & $ 2.67 \pm 0.18 $ & $ 1.79 \pm 0.22 $ & $ 1.12 \pm 0.21 $ & $ 0.38 \pm 0.08 $ & $ 2.57 \pm 0.15 $ \\ \hline
Hopper 2 $(\times 10^3)$ & $ \bm{2.51 \pm 0.07} $ & $ 2.26 \pm 0.12 $ & $ 1.49 \pm 0.15 $ & $ 1.15 \pm 0.11 $ & $ 0.66 \pm 0.13 $ & $ 1.89 \pm 0.08 $ \\ \hline
Hopper 3 $(\times 10^3)$ & $ 0.85 \pm 0.07 $ & $ 0.79 \pm 0.04 $ & $ 0.87 \pm 0.08 $ & $ 0.49 \pm 0.11 $ & $ 0.47 \pm 0.08 $ & $ \bm{1.50 \pm 0.07} $ \\
\toprule\bottomrule
HumanoidStandup 1 $(\times 10^5)$ & $ 1.08 \pm 0.04 $ & $ 1.03 \pm 0.07 $ & $ 1.05 \pm 0.06 $ & $ 0.99 \pm 0.06 $ & $ 0.77 \pm 0.06 $ & $ \bm{1.12 \pm 0.05} $ \\ \hline
HumanoidStandup 2 $(\times 10^5)$ & $ 0.97 \pm 0.04 $ & $ \bm{1.07 \pm 0.05} $ & $ 0.93 \pm 0.04 $ & $ 0.92 \pm 0.04 $ & $ 0.85 \pm 0.08 $ & $ 1.06 \pm 0.04 $ \\ \hline
HumanoidStandup 3 $(\times 10^5)$ & $ \bm{1.09 \pm 0.06} $ & $ 1.04 \pm 0.03 $ & $ 0.98 \pm 0.06 $ & $ 1.01 \pm 0.04 $ & $ 0.87 \pm 0.07 $ & $ 1.04 \pm 0.07 $ \\
\toprule\bottomrule
InvertedPendulum 1 $(\times 10^2)$ & $ \bm{9.66 \pm 0.25} $ & $ 9.17 \pm 0.54 $ & $ 8.79 \pm 0.87 $ & $ 4.51 \pm 0.12 $ & $ 4.21 \pm 0.81 $ & $ 9.32 \pm 0.11 $ \\ \hline
InvertedPendulum 2 $(\times 10^2)$ & $ 6.13 \pm 1.42 $ & $ 6.26 \pm 0.95 $ & $ 8.27 \pm 0.70 $ & $ 1.76 \pm 0.51 $ & $ 3.07 \pm 0.66 $ & $ \bm{9.18 \pm 0.07} $ \\ \hline
\toprule\bottomrule
Walker 1 $(\times 10^3)$ & $ \bm{3.70 \pm 0.31} $ & $ 3.51 \pm 0.16 $ & $ 2.03 \pm 0.26 $ & $ 1.55 \pm 0.25 $ & $ 0.15 \pm 0.03 $ & $ 3.59 \pm 0.26 $ \\ \hline
Walker 2 $(\times 10^3)$ & $ \bm{4.72 \pm 0.12} $ & $ 4.37 \pm 0.32 $ & $ 2.39 \pm 0.20 $ & $ 1.63 \pm 0.22 $ & $ 0.26 \pm 0.05 $ & $ 4.54 \pm 0.31 $ \\ \hline
Walker 3 $(\times 10^3)$ & $ 4.27 \pm 0.21 $ & $ 4.21 \pm 0.30 $ & $ 2.48 \pm 0.24 $ & $ 1.65 \pm 0.15 $ & $ 0.21 \pm 0.07 $ & $ \bm{4.48 \pm 0.16} $ \\
\toprule\bottomrule
Small HalfCheetah 1 $(\times 10^3)$ & $ 6.00 \pm 0.15 $ & $ 6.04 \pm 0.13 $ & $ 5.71 \pm 0.07 $ & $ 3.38 \pm 0.34 $ & $ -0.42 \pm 0.01 $ & $ \bm{8.11 \pm 0.17} $ \\ \hline
Small Hopper 1 $(\times 10^3)$ & $ 2.96 \pm 0.31 $ & $ 2.44 \pm 0.31 $ & $ 1.57 \pm 0.21 $ & $ 1.53 \pm 0.21 $ & $ 0.56 \pm 0.13 $ & $ \bm{3.44 \pm 0.10} $ \\
\toprule
\end{tabular}
}
\end{center}
\end{table*}

\begin{table}[!t]
\centering
\caption{Avg.\ $\pm$ std.\ error of worst-case performance $R_\text{worst}(\mu)$ and average performance $R_\text{average}(\mu)$ on InvertedPendulum 1 obtained by \proposed{} variants over 10 trials}
\label{table: ablationstudy_m2td3}
\fontsize{8}{10}\selectfont
\begin{tabular}{|c|c|c|c|c|c|}
\bottomrule
Environment & N=5 & N=1 & N=10 & w/o DRS & w/o PRS \\
\toprule\bottomrule
worst-case $(\times 10^2)$ & $ 8.22 \pm 1.13 $ & $ 5.77 \pm 1.29 $ & $ 9.07 \pm 0.89 $ & $ 9.07 \pm 0.88 $ & $ 5.41 \pm 1.45 $ \\ \hline
average $(\times 10^2)$ & $ 9.66 \pm 0.25 $ & $ 8.10 \pm 0.90 $ & $ 9.89 \pm 0.11 $ & $ 9.08 \pm 0.88 $ & $ 6.90 \pm 1.25 $ \\
\toprule
\end{tabular}
\end{table}

\paragraph{Ablation Study}
Ablation studies were conducted on InvertedPendulum 1 with the pole mass as the uncertainty parameters.
\proposed{} with $N = 5$ was taken as the baseline.
We tested several variants as follows: with different multiple uncertainty parameters ($N=1$ and $10$), without a distance-based refreshing strategy (w/o DRS), and without a probability-based refreshing strategy (w/o PRS).
\Cref{table: ablationstudy_m2td3} shows the results.
The inclusion of multiple uncertainty parameters and the probability-based refresh strategy contributed significantly to the worst-case performance and average performance, implying that both techniques contribute to better estimating the worst uncertainty parameter.
Although DRS had little impact on the performance, prior knowledge in DRS can be implemented by defining the distance and the threshold in the uncertainty parameter task-dependently, which we did not implement here.

\paragraph{Small Uncertainty Set (Limitation of \proposed)}
Small HalfCheetah 1 and Small Hopper 1 were designed to have a smaller uncertainty parameter set than HalfCheetah 1 and Hopper 1 and to reveal the effect of the size of the uncertainty parameter interval.
As expected, a smaller uncertainty parameter set resulted in higher worst-case (\Cref{table:results_mujoco}) and average (\Cref{table:results_mujoco_mean}) performances, for all approaches.
Because \proposed{} performs the worst-case optimization, it is expected to show better worst-case performance than DR, independently of the size of the uncertainty parameter set.
However, in these senarios, DR showed better worst-case performance than that of \proposed{}, while \proposed{} achieved competitive or superior worst-case performance to those of DR on HalfCheetah 1 and Hopper 1.
This may be because the max-min optimization performed by \proposed{} resulted in sub-optimal policies.
When the uncertainty parameter set is small, and the performance does not change significantly over the uncertainty parameter set, maximizing the average performance is likely to lead to high worst-case performance.
The sub-optimal results obtained by \proposed{} is then dominated by the results obtained by DR.
Therefore, the maxmin optimization in \proposed{} can be improed.

\paragraph{Evaluation Under Adversarial External Force}
Although we developed the proposed approach for the situation that the uncertainty parameter is directly encoded by $\omega$, we can extend it to the situation where the model misspecification is expressed by an external force produced by an adversarial agent as in \cite{DBLP:conf/icml/PintoDSG17}.
The extended approach and its experimental result are given in \Cref{apdx:evaluate_under_adv-ex-force}.
Superior worst-case performances of \proposed{} over DR, RARL, and M3DDPG were observed for this setting as well.

\section{Conclusion} \label{sec:conclusion}

In this study, we targeted the policy optimization aimed at maximizing the worst-case performance in a predefined uncertainty set.
The list of the contributions are as follows.
(i) We formulated the off-policy deterministic actor-critic approach to the worst-case performance maximization as a tri-level optimization problem \eqref{eq:tri}.
(ii) We developed the worst-case performance of \proposed. The key concepts were the incorporation of the uncertainty parameter into the critic network and use of the simultaneous gradient ascent descent method. Different technical components were introduced to stabilize the training.
(iii) We evaluated the worst-case performance of \proposed{} on 19 MuJoCo tasks through comparison with three baseline methods. Ablation studies revealed the usefulness of each component of \proposed.

\section*{Acknowledgements}

This research is partially supported by the JSPS KAKENHI Grant Number 19H04179 and the NEDO Project Number JPNP18002.

\bibliography{ref}
\bibliographystyle{plain}

\section*{Checklist}

\begin{enumerate}

\item For all authors...
\begin{enumerate}
  \item Do the main claims made in the abstract and introduction accurately reflect the paper's contributions and scope?
    \answerYes{Please see \Cref{sec:proposed} and \Cref{sec:ex}.}
  \item Did you describe the limitations of your work?
    \answerYes{Please see paragraph titled "Small Uncertainty Set" in \Cref{sec:ex} and \Cref{sec:conclusion}.}
  \item Did you discuss any potential negative societal impacts of your work?
    \answerNA{}
  \item Have you read the ethics review guidelines and ensured that your paper conforms to them?
    \answerYes{}
\end{enumerate}

\item If you are including theoretical results...
\begin{enumerate}
  \item Did you state the full set of assumptions of all theoretical results?
    \answerNA{}
        \item Did you include complete proofs of all theoretical results?
    \answerNA{}
\end{enumerate}

\item If you ran experiments...
\begin{enumerate}
  \item Did you include the code, data, and instructions needed to reproduce the main experimental results (either in the supplemental material or as a URL)?
    \answerYes{Please see \Cref{apdx:algo}.}
  \item Did you specify all the training details (e.g., data splits, hyperparameters, how they were chosen)?
    \answerYes{Please see \Cref{sec:ex} and \Cref{apdx:ex}.}
        \item Did you report error bars (e.g., w.r.t.~the random seed after running experiments multiple times)?
    \answerYes{}
        \item Did you include the total amount of compute and the type of resources used (e.g., type of GPUs, internal cluster, or cloud provider)?
    \answerYes{Please see \Cref{apdx:ex}.}
\end{enumerate}

\item If you are using existing assets (e.g., code, data, models) or curating/releasing new assets...
\begin{enumerate}
  \item If your work uses existing assets, did you cite the creators?
    \answerYes{}
  \item Did you mention the license of the assets?
    \answerYes{Code licence is part of the repository.}
  \item Did you include any new assets either in the supplemental material or as a URL?
    \answerYes{We include our code in the \Cref{apdx:algo}.}
  \item Did you discuss whether and how consent was obtained from people whose data you're using/curating?
    \answerNA{}
  \item Did you discuss whether the data you are using/curating contains personally identifiable information or offensive content?
    \answerNA{}
\end{enumerate}

\item If you used crowdsourcing or conducted research with human subjects...
\begin{enumerate}
  \item Did you include the full text of instructions given to participants and screenshots, if applicable?
    \answerNA{}
  \item Did you describe any potential participant risks, with links to Institutional Review Board (IRB) approvals, if applicable?
    \answerNA{}
  \item Did you include the estimated hourly wage paid to participants and the total amount spent on participant compensation?
    \answerNA{}
\end{enumerate}

\end{enumerate}


\appendix

\section{Algorithm Details}\label{apdx:algo}
\Cref{alg:proposed} summarizes the overall framework of \proposed{}.
The training is stalled if the size of the replay buffer is smaller than the minibatch size, i.e., if $\abs{B} < M$.
\Cref{alg:critic,alg:actor} show the critic network update and the actor network and uncertainty parameter sampler update, respectively.
Although we write the gradient-based update in the form of a mini-batch stochastic gradient update for simplicity, we employ an adaptive approach such as Adam~\cite{DBLP:journals/corr/KingmaB14}.

\begin{algorithm}[ht]
  \caption{\proposed{} (Framework)}
  \label{alg:proposed}
  \begin{algorithmic}[1]
  \renewcommand{\algorithmicrequire}{\textbf{Input:}}
  \renewcommand{\algorithmicensure}{\textbf{Output:}}
  \STATE \texttt{\# initialization}
  \STATE Initialize uncertainty parameters $\hat{\omega}_i \in \mathcal{U}(\Omega)$ for $i=1,\dots,N$
  \STATE Initialize policy parameter $\theta$ and critic parameters $\phi_1$, $\phi_2$ with random values
  \STATE Initialize target network parameters $\theta' \leftarrow \theta$, $\phi_1' \leftarrow \phi_1$, $\phi_2' \leftarrow \phi_2$
  \STATE Initialize frequency parameter $p_1,\dots,p_N = 1/N$
  \STATE Initialize replay buffer $B = \emptyset$
  \STATE \texttt{\# training loop}
  \STATE Draw uncertainty parameter $\omega \sim \alpha_0$
  \STATE Observe initial state $s \sim p_\omega^0$
  \FOR {$\mathrm{t}=1$ to $T_\text{max}$}
    \STATE \texttt{\# interaction}
    \STATE Select action $a \sim \beta_t(s)$
    \STATE Interact with $\mathcal{M}_\omega$ with $a$, observe next state $s'$, immediate reward $r$ and termination flag $h$
    \STATE Store transition tuple $(s, a, r, s', h, \omega)$ in $B$
    \IF {$h = 1$}
        \STATE Reset uncertainty parameter $\omega \sim \alpha_t$
        \STATE Observe initial state $s \sim p_\omega^0$
    \ELSE
        \STATE Update current state $s \leftarrow s'$
    \ENDIF
    \STATE \texttt{\# learning}
    \STATE Sample mini-batch $\{(s_i, a_i, r_i, s_i', h_i, \omega_i)\}$ of $\mathrm{M}$ transitions uniform-randomly from $B$
    \STATE Perform \Cref{alg:critic} for critic network update
    \IF {$\mathrm{mod}(t, T_\text{freq}) = 0$}
        \STATE Perform \Cref{alg:actor} for actor network update and uncertainty parameter sampler update
        \STATE Update target networks analogously to TD3
    \ENDIF
  \ENDFOR
  \end{algorithmic}
\end{algorithm}

\begin{algorithm}[ht]
  \caption{Critic Update}
  \label{alg:critic}
  \begin{algorithmic}[1]
  \renewcommand{\algorithmicrequire}{\textbf{Input:}}
  \renewcommand{\algorithmicensure}{\textbf{Output:}}
  \FOR{$i = 1,\dots,M$}
    \STATE $\tilde{a}_i \leftarrow \mu_{\theta'}(s_i') + \epsilon_a$, $\epsilon_a \sim \Pi_{a}(\mathcal{N}(0, \tilde{\Sigma}_a))$
    \STATE $\tilde{\omega}_i' \leftarrow \omega_i + \epsilon_\omega$, $\epsilon_\omega \sim \Pi_{\omega}(\mathcal{N}(0, \tilde{\Sigma}_\omega))$
    \STATE $y_i \leftarrow r_i + \min\{ Q_{\phi'_1}(s_i', \tilde{a}_i, \tilde{\omega}_i), Q_{\phi'_2}(s_i', \tilde{a}_i, \tilde{\omega}_i)\}$
  \ENDFOR
  \FOR{$j=1,2$}
    \STATE $\phi_j \leftarrow \phi_j - \lambda_{\phi} \nabla_{\phi_j} \frac{1}{M} \sum_{i=1}^{M} (y_i - Q_{\phi_j}(s_i, a_i, \omega_i))^2$
  \ENDFOR
  \end{algorithmic}
\end{algorithm}

\begin{algorithm}[ht]
  \caption{Actor Update and Uncertainty Parameter Sampler Update}
  \label{alg:actor}
  \begin{algorithmic}[1]
  \renewcommand{\algorithmicrequire}{\textbf{Input:}}
  \renewcommand{\algorithmicensure}{\textbf{Output:}}
  \STATE \texttt{\# maximin update}
  \STATE $k' = \argmin_{k} \frac{1}{M}\sum_{i=1}^{M} Q_{\phi^1}(s_i, \mu_{\theta}(s_i), \hat{\omega}_k)$
  \STATE $\theta \leftarrow \theta + \lambda_\theta \nabla_\theta \frac{1}{M}\sum_{i=1}^{M} Q_{\phi^1}(s_i, \mu_{\theta}(s_i), \hat{\omega}_{k'})$
  \STATE $\hat{\omega}_{k'} \leftarrow \hat{\omega}_{k'} - \lambda_\omega \nabla_{\hat{\omega}_{k'}} \frac{1}{M}\sum_{i=1}^{M} Q_{\phi^1}(s_i, \mu_{\theta}(s_i), \hat{\omega}_{k'})$
  \STATE \texttt{\# uncertainty parameter sampler update with refreshing strategy}
  \FOR{$k=1,\dots,N$}
    \STATE $\hat{\omega}_k \sim \mathcal{U}(\Omega)$ \textbf{if} $d_{\omega}(\hat{\omega}_k, \hat{\omega}_\ell) \leq d_\mathrm{thre}$ for some $\ell \neq k$
    \STATE $\hat{\omega}_k \sim \mathcal{U}(\Omega)$ \textbf{if} $p_k \leq p_\mathrm{thre}$
    \IF{$\hat{\omega}_k$ is refreshed}
      \STATE $p_k \leftarrow 1/N$
    \ELSE
      \STATE $p_k \leftarrow (1 - 1/T_\mathrm{last}) p_k + (1/T_\mathrm{last}) \mathbb{I}\{k = k'\}$
    \ENDIF
  \ENDFOR
  \STATE $p_k \leftarrow p_k / \sum_{\ell = 1}^{n} p_{\ell}$ for all $k = 1,\dots,N$
  \end{algorithmic}
\end{algorithm}

We maintain the frequency $p_k$ of each uncertainty parameter $\hat{\omega}_k$ being the worst one among $\hat{\omega}_1,\dots,\hat{\omega}_N$ in \Cref{alg:actor}. This is used in two ways: criteria for the refreshing strategy of $\hat{\omega}_k$ in \Cref{alg:actor}; and mixture weights for the uncertainty parameter sampler $\alpha$.
The update of $p_k$ follows the exponential moving average with the momentum $(1/T_\text{last})$, where $T_\text{last}$ is the number of steps spent in the last episode ($T_\text{last}$ is set to $1000$ for the first episode). The reason behind this design choice is as follows.
The short episode is a meaning that a bad uncertainty parameter $\omega$ is used in the last episode.
Because the uncertainty parameter $\omega$ used in the interaction is sampled from $\alpha$, which is a gaussian mixture with components centered at $\hat{\omega}_1,\dots,\hat{\omega}_k$, it implies that they include a bad uncertainty parameter as well.
Then, there is a high chance that this uncertainty parameter is selected as the worst uncertainty parameter $\hat{\omega}_{k'}$ in \Cref{alg:actor}.
By setting a greater momentum $(1/T_\text{last})$, we can accelerate the approach of $p_k$ to $\mathbb{I}\{k = k'\}$, which is $1$ if $k = k'$ and $0$ otherwise.
With this fast update of $p_k$, we expect two consequences.
First, there are increased chances to sample $\omega$ around the worst $\hat{\omega}_{k'}$.
Second, there are increased chances to refresh the non-worst uncertainty parameters, which leads to more exploration of the worst uncertainty parameter search.
Preliminary experiments have confirmed that this momentum setting leads to a better worst-case performance than a constant momentum.

Our implementation is publicly available (\url{https://github.com/akimotolab/M2TD3}).

\section{Soft-Min Variant: SoftM2TD3} \label{apdx:softm2td3}
In theory, \proposed{} can obtain a policy that exhibits a better worst-case performance than a policy obtained by DR as DR does not explicitly maximize the worst-case performance.
However, we empirically observe that DR sometimes achieves a better worst-case performance than \proposed{} on senarios where the performance does not change significantly over the uncertainty set $\Omega$ (\Cref{table:results_mujoco}).
We conjecture that this is because the difficulty in the max-min optimization of \proposed{} compared to the optimization of the expectation in DR.

To mitigate this issue, we propose a variant of \proposed{}, called Soft\proposed{}.
The objective of the update of the policy parameter $\theta$ is replaced with the following soft-min version:
\begin{align}
        \tilde{J}_t(\theta) &= \sum_{k=1}^N w_k \left[  \frac{1}{M}\sum_{i=1}^{M} Q_{\phi_t}(s_i, \mu_{\theta}(s), \hat{\omega}_{k}) \right]\enspace, \label{eq: softm2td3}
\end{align}
where $w_k$ is the weight for uncertainty parameter $\hat{\omega}_k$. A greater weight value should be assigned to $\hat{\omega}_k$ with smaller Q-values. We used the frequency $p_k$ of $\hat{\omega}_k$ as the worst-case during the actor update, i.e., $w_k = p_k$. This update is close to \proposed{} if $p = (p_1, \dots, p_N)$ is close to a one-hot vector and is close to DR if $p_{1} \approx \cdots \approx p_N$, which is the case at the beginning of the training.

The objective function of SoftM2TD3 is considered an approximation of the objective function of \proposed{}.
The difference between \proposed{} and SoftM2TD3 is in the optimization process.
Because the objective function of SoftM2TD3 is closer to that of DR, we expect that the optimization in SoftM2TD3 is more efficient than \proposed{}, where the effect of the accuracy of the estimation of the worst-case uncertainty parameters is greater than SoftM2TD3.

\section{Experiment Details}\label{apdx:ex}

\begin{table*}[t]
\centering
\caption{List of senarios used in the experiments}
\label{table:environment_settings}
\fontsize{6}{10}\selectfont
\begin{tabular}{|l|l|l|l|}
\bottomrule
Environment          & Uncertainty Set $\Omega$                         & Reference Parameter & Uncertainty Parameter Name                                 \\ \toprule\bottomrule
\multicolumn{4}{|l|}{Baseline MuJoCo Environment: Ant}\\ \hline
Ant 1                 & {[}0.1, 3.0{]}                                    & 0.33                & torso mass
 \\ \hline
Ant 2                 & {[}0.1, 3.0{]} $\times$ {[}0.01, 3.0{]}                 & (0.33, 0.04)         & torso mass $\times$ front left leg mass
 \\ \hline
Ant 3                 & {[}0.1, 3.0{]} $\times$ {[}0.01, 3.0{]} $\times$ {[}0.01, 3.0{]} & (0.33, 0.04, 0.06) & torso mass $\times$ front left leg mass $\times$ front right leg mass
 \\ \toprule \bottomrule
\multicolumn{4}{|l|}{Baseline MuJoCo Environment: HalfCheetah}\\ \hline
HalfCheetah 1        & {[}0.1, 4.0{]}                                    & 0.4                 & world friction                                 \\ \hline
HalfCheetah 2        & {[}0.1, 4.0{]} $\times$ {[}0.1, 7.0{]}                  & (0.4, 6.36)          & world friction $\times$ torso mass                    \\ \hline
HalfCheetah 3        & {[}0.1, 4.0{]} $\times$ {[}0.1, 7.0{]} $\times$ {[}0.1, 3.0{]}  & (0.4, 6.36, 1.53)   & world friction $\times$ torso mass $\times$ back thigh mass  \\ \toprule\bottomrule
\multicolumn{4}{|l|}{Baseline MuJoCo Environment: Hopper}\\ \hline
Hopper 1              & {[}0.1, 3.0{]}                                    & 1.00                & world friction                                 \\ \hline
Hopper 2              & {[}0.1, 3.0{]} $\times$ {[}0.1, 3.0{]}                   & (1.00, 3.53)         & world friction $\times$ torso mass                    \\ \hline
Hopper 3              & {[}0.1, 3.0{]} $\times$ {[}0.1, 3.0{]} $\times$ {[}0.1, 4.0{]}  & (1.00, 3.53, 3.93)  & world friction $\times$ torso mass $\times$ thigh mass       \\ \toprule\bottomrule
\multicolumn{4}{|l|}{Baseline MuJoCo Environment: HumanoidStandup}\\ \hline
HumanoidStandup 1    & {[}0.1, 16.0{]}                                   & 8.32                & torso mass                                     \\ \hline
HumanoidStandup 2    & {[}0.1, 16.0{]} $\times$ {[}0.1, 8.0{]}                  & (8.32, 1.77)         & torso mass $\times$ right foot mass                   \\ \hline
HumanoidStandup 3    & {[}0.1, 16.0{]} $\times$ {[}0.1, 5.0{]} $\times$ {[}0.1, 8.0{]} & (8.32, 1.77, 4.53)  & torso mass $\times$ right foot mass $\times$ left thigh mass \\ \toprule\bottomrule
\multicolumn{4}{|l|}{Baseline MuJoCo Environment: Inveted Pendulum}\\ \hline
InvertedPendulum 1   & {[}1.0, 31.0{]}                                   & 4.90                & pole mass                                      \\ \hline
InvertedPendulum 2   & {[}1.0, 31.0{]} $\times$ {[}1.0, 11.0{]}                                               & (4.90, 9.42)         & pole mass $\times$ cart mass                          \\ \toprule\bottomrule
\multicolumn{4}{|l|}{Baseline MuJoCo Environment: Walker}\\ \hline
Walker 1             & {[}0.1, 4.0{]}                                    & 0.7                & world friction                                 \\ \hline
Walker 2             & {[}0.1, 4.0{]} $\times$ {[}0.1, 5.0{]}                   & (0.7, 3.53)          & world friction $\times$ torso mass                    \\ \hline
Walker 3             & {[}0.1, 4.0{]} $\times$ {[}0.1, 5.0{]} $\times$ {[}0.1, 6.0{]}  & (0.7, 3.53, 3.93)& world friction $\times$ torso mass $\times$ thigh mass       \\ \toprule \bottomrule
\multicolumn{4}{|l|}{Small MuJoCo Environments}\\ \hline
Small HalfCheetah 1             & {[}0.1, 3.0{]}                                    & 0.4                & world friction                                 \\ \hline
Small Hopper 1              & {[}0.1, 2.0{]}                                    & 1.00                & world friction                                              \\ \toprule
\end{tabular}
\end{table*}

\Cref{table:environment_settings} lists the senario we used in our experiments.
These senarios are created by setting 1 to 3 constants in the original MuJoCo environment to the uncertainty parameters.
The uncertainty set is designed as an interval.
Except for Hopper 2 and Hopper 3, the original value is included in the uncertainty set; hence, the trivial upper bound of the worst-case performance is the best performance of the corresponding MuJoCo environment.

In all approaches in all senarios, the same configurations are used for fair comparison as follows.

For DR and RARL, the input to the critic is a $(s, a)$ pair, whereas it is a tuple of $(s, a, \omega)$ for \proposed{}, SoftM2TD3, and M3DDPG.
Except for this point, we used the same network architecture. The policy and the critic networks are defined as fully connected layers with two hidden layers of size 256.

The uncertainty parameter sampler $\alpha_t$ is
\begin{equation}
    \alpha_t = \begin{cases} \mathcal{U}(\Omega) & t \leq T_\text{rand} \\ \Pi_\Omega(\sum_{k=1}^{N} p_k \cdot \mathcal{N}(\hat{\omega}_k, \Sigma_\omega)) & t > T_\text{rand} \enspace,\end{cases}
\end{equation}
where $\Sigma_\omega$ is a diagonal matrix, with diagonal elements 0.5 times the lengths of the intervals of the corresponding dimension of $\Omega$, and $\Pi_\Omega$ is the projection onto $\Omega$ and $T_\text{rand} = 10^5$.
The behavior policy $\beta_t$ is
\begin{equation}
    \beta_t = \begin{cases} \mathcal{U}(A) & t \leq T_\text{rand} \\ \Pi_{A}(\mathcal{N}(\mu_\theta(s), \Sigma_{a})) & t > T_\text{rand} \enspace,\end{cases}
\end{equation}
where $\Sigma_a$ is a diagonal matrix, with diagonal elements are the 0.5 times the length of the intervals of the corresponding dimension of $A$, and $\Pi_A$ is the projection onto $A$.
The $\Sigma_\omega$ element is designed to decay at each time step to 0.05 times the length of the intervals of it when the learning progresses to half of the total.

The minibatch size is $M = 100$.
The learning rates for the actor update, the uncertainty parameter update and the critic update are $\lambda_\theta = \lambda_\omega = \lambda_\phi = 3 \times 10^{-4}$.
The noise covariance matrices for the target policy smoothing are $\tilde{\Sigma}_a = 2 \Sigma_a$ and $\tilde{\Sigma}_{\omega} = 2 \Sigma_\omega$.
The noise for the target policy smoothing is clipped by $\Pi_a$ and $\Pi_\omega$ into the ranges of $\pm 0.25$ times the interval lengths of the corresponding dimensions of $A$ and $\Omega$, respectively.
The actor and target network update frequency is $T_\text{freq} = 2$.
The above parameter settings follow TD3 \cite{fujimoto2018addressing}.

We used $N = 5$ uncertainty parameters.
For the refreshing strategy of the uncertainty parameters, we used $\ell_1$-distance as $d_\omega$.
The distance threshold is $d_\mathrm{thre} = 0.1$.
The frequency threshold is $0.05$.

Experiments were performed on a machine with two NVIDIA RTX A5000 GPUs, two Intel(R) Xeon(R) Gold 6230 CPUs, and 192GB memory.

\section{Related Work}\label{apdx:related}

\Cref{table:robust_rrl} summarizes the robust policy search method taxonomy.
\Cref{table:def_q} compares the related methods considering their action value function definitions.

\begin{table}[t]
  \centering
  \caption{Taxonomy of robust policy search. Cont.: Continuous space. Disc.: Discrete space.}\label{table:robust_rrl}%
  \fontsize{8}{10}\selectfont%
  \begin{tabular}{|l||c|c|c|c|c|}
  \bottomrule
  Method & Uncertainty  & $S$ \& $A$ & $\Omega$ \\ \toprule\bottomrule
  R-MPO & Transition & Cont. & Disc.\\ \hline
  ROPI & Transition & Disc. & Cont. \\ \hline
  ATLA, SA-DRL & Observation & Cont. & Cont. \\ \hline
  \makecell[l]{PAIRED,  RARL,  M3DDPG\\ RRL, Adv-SAC, RSAC, \proposed{}} & \makecell{Reward \& \\ Transition} & Cont. & Cont.\\ \toprule
  \end{tabular}
\end{table}

\begin{table}[t]
 \centering
 \caption{Comparison of action value functions in related methods}\label{table:def_q}%
 \fontsize{8}{10}\selectfont%
 \begin{tabular}[\hsize]{|l||l|}
 \bottomrule
 Method & Action Value Function\\ \toprule\bottomrule
 R-MPO, ROPI, Robust DP & $Q(s, a) = r(s, a) + \gamma \min_{\omega \in \Omega} \mathbb{E}[Q(s', \mu_{\theta}(s')) \mid s' \sim p_{\omega}(\cdot \mid s, a)]$ \\ \hline
 DR& $Q(s, a) = \mathbb{E}_{\omega \in \Omega} [r_\omega(s, a) + \gamma  \mathbb{E}[Q(s', \mu_{\theta}(s')) \mid s' \sim p_\omega(\cdot \mid s, a)]]$\\ \hline
 RRL, RARL, Adv-SAC & $Q_\omega(s, a) =  r_\omega(s, a) + \gamma  \mathbb{E}[Q_\omega(s', \mu_{\theta}(s')) \mid s' \sim p_\omega(\cdot \mid s, a)]$ \\ \hline
 M3DDPG & $Q(s, a, \omega) = r_\omega(s, a) + \gamma \mathbb{E}[\min_{\omega' \in \Omega} Q(s', \mu_{\theta}(s'), \omega') \mid s' \sim p_{\omega}(\cdot \mid s, a)]$ \\ \hline
 \proposed{} & $Q(s, a, \omega) =  r_\omega(s, a) + \gamma \mathbb{E}[Q(s', \mu_\theta(s'), \omega) \mid s' \sim p_\omega(\cdot \mid s, a)]$ \\ \toprule
 \end{tabular}
\end{table}

\section{Performance of TD3}\label{apdx:td3}

\begin{table}[t]
\begin{center}
\caption{Avg. $\pm$ std. error of worst-case performance $R_\text{worst}(\mu)$ and average performance $R_\text{average}(\mu)$ and reference performance $R_\text{ref}(\mu)$ over 10 trials for TD3 (reference parameter)}
\label{table:td3_reference}
\scalebox{0.7}{
\begin{tabular}{|c|c|c|}
\bottomrule
Environment           & worst & average         \\ 
\toprule\bottomrule
\multicolumn{3}{|l|}{Ant (reference parameters): $3.02 \pm 0.15 (\times 10^3)$} \\ \hline
Ant 1 $(\times 10^3)$                &$2.22 \pm 0.50 $& $2.76 \pm 0.50$ \\ \hline
Ant 2 $(\times 10^3)$              &$1.59 \pm 0.08$& $2.28 \pm 0.09$  \\ \hline
Ant 3 $(\times 10^2)$                &$-0.99 \pm 1.13$& $3.16 \pm 1.00$ \\
\toprule\bottomrule
\multicolumn{3}{|l|}{HalfCheetah (reference parameters): $10.2 \pm 0.2 (\times 10^3)$} \\ \hline
Small HalfCheetah 1 $(\times 10^3)$ & $0.03 \pm 0.11$ & $3.73 \pm 0.29$ \\ \hline
HalfCheetah 1 $(\times 10^3)$       &$-0.34 \pm 0.04$ & $2.79 \pm 0.22$ \\ \hline
HalfCheetah 2 $(\times 10^3)$       &$-0.53 \pm 0.06$ & $2.63 \pm 0.20$  \\ \hline
HalfCheetah 3 $(\times 10^3)$       &$-0.61 \pm 0.08$ & $2.47 \pm 0.18$  \\
\toprule\bottomrule
\multicolumn{3}{|l|}{Hopper (reference parameters): $3.01 \pm 0.19 (\times 10^3)$} \\ \hline
Small Hopper 1 $(\times 10^3)$ & $2.84 \pm 0.22$ & $2.97 \pm 0.21$\\ \hline
Hopper 1 $(\times 10^3)$        &$0.40 \pm 0.02$ & $2.39 \pm 0.14$\\ \hline
Hopper 2 $(\times 10^3)$        &$0.21 \pm 0.04$ & $1.54 \pm 0.17$ \\ \hline
Hopper 3 $(\times 10^3)$        &$0.14 \pm 0.03$ & $1.15 \pm 0.14$\\ 
\toprule\bottomrule
\multicolumn{3}{|l|}{HumanoidStandup (reference parameters): $1.08 \pm 0.03 (\times 10^5)$} \\ \hline
HumanoidStandup 1 $(\times 10^{5})$   &$0.85 \pm 0.07$  & $1.03 \pm 0.04$\\ \hline
HumanoidStandup 2 $(\times 10^{5})$   &$0.73 \pm 0.07$ & $1.03 \pm 0.03$\\ \hline
HumanoidStandup 3 $(\times 10^{5})$   &$0.57 \pm 0.04$ & $1.01 \pm 0.03$\\
\toprule\bottomrule
\multicolumn{3}{|l|}{InvertedPendulum (reference parameters): $10.0 \pm 0.0 (\times 10^2)$} \\ \hline
InvertedPendulum 1 $(\times 10^{2})$   &$0.24 \pm 0.10$ & $7.34 \pm 0.76$\\ \hline
InvertedPendulum 2 $(\times 10^{2})$   &$0.03 \pm 0.00$ & $4.05 \pm 0.52$\\
\toprule\bottomrule
\multicolumn{3}{|l|}{Walker (reference parameters): $4.08 \pm 0.16 (\times 10^3)$} \\ \hline
Walker 1 $(\times 10^{3})$  &$0.68 \pm 0.12$ & $3.12 \pm 0.20$\\ \hline
Walker 2 $(\times 10^{3})$  &$0.28 \pm 0.07$ & $2.70 \pm 0.20$ \\ \hline
Walker 3 $(\times 10^{3})$  &$0.17 \pm 0.06$ & $2.60 \pm 0.18$\\
\toprule
\end{tabular}
}
\end{center}
\end{table}

\Cref{table:td3_reference} summarizes the worst-case performance and the average performance of the policies obtained by TD3 on the Ant, HalfCheetah, Hopper, HumanoidStandup, InvertedPendulum, and Walker environments with their reference parameters.
The results show that the original TD3 policies learned under the reference parameter cannot be generalized to uncertain parameters set in most scenarios. These results provide the baseline performances on these environments and show that these environments are non-trivial for the worst-case and the average performance maximization.

\section{Original RARL} \label{apdx:RARL_TRPO}
The performance of the original RARL, whose baseline RL approach is TRPO, is shown in \Cref{table:results_mujoco_trpo_rarl}.
RARL (TRPO) exhibited low worst-case performances, similarly to those of RARL (DDPG).
These low performances of RARL (TRPO) may be due to its defect of the optimization strategy and because TRPO is an on-policy method, which often requires a greater number of interactions than off-policy methods~\cite{DBLP:conf/icml/HaarnojaZAL18}.

\begin{table}[t]
\begin{center}
\caption{Avg. $\pm$ std. error of worst-case performance $R_\text{worst}(\mu)$ and average performance $R_\text{average}(\mu)$ over 10 trials for RARL (TRPO)}
\label{table:results_mujoco_trpo_rarl}
\scalebox{0.7}{
\begin{tabular}{|c|c|c|}
\bottomrule
Environment           & worst & average            \\ 
\toprule\bottomrule
Ant 1 $(\times 10^1)$ & $ -4.92 \pm 0.48 $ & $ -2.29 \pm 0.22 $ \\ \hline
Ant 2 $(\times 10^2)$ & $ -1.15 \pm 0.19 $ & $ -0.37 \pm 0.04 $ \\ \hline
Ant 3 $(\times 10^2)$ & $ -0.32 \pm 0.41 $ & $ 1.28 \pm 0.68 $ \\
\toprule\bottomrule
HalfCheetah 1 $(\times 10^2)$ & $ -3.06 \pm 0.85 $ & $ 1.22 \pm 0.63 $ \\ \hline
HalfCheetah 2 $(\times 10^2)$ & $ -5.23 \pm 0.89 $ & $ -0.03 \pm 0.73 $ \\ \hline
HalfCheetah 3 $(\times 10^2)$ & $ -9.70 \pm 1.89 $ & $ -0.70 \pm 0.65 $ \\
\toprule\bottomrule
Hopper 1 $(\times 10^2)$ & $ 2.89 \pm 0.25 $ & $ 3.49 \pm 0.38 $ \\ \hline
Hopper 2 $(\times 10^2)$ & $ 2.67 \pm 0.26 $ & $ 4.82 \pm 0.64 $ \\ \hline
Hopper 3 $(\times 10^2)$ & $ 0.71 \pm 0.13 $ & $ 2.24 \pm 0.47 $ \\
\toprule\bottomrule
HumanoidStandup 1 $(\times 10^4)$ & $ 5.30 \pm 0.22 $ & $ 6.78 \pm 0.21 $ \\ \hline
HumanoidStandup 2 $(\times 10^4)$ & $ 5.04 \pm 0.09 $ & $ 6.99 \pm 0.25 $ \\ \hline
HumanoidStandup 3 $(\times 10^4)$ & $ 4.99 \pm 0.08 $ & $ 6.64 \pm 0.17 $ \\
\toprule\bottomrule
InvertedPendulum 1 $(\times 10^2)$ & $ 0.44 \pm 0.10 $ & $ 3.03 \pm 0.98 $ \\ \hline
InvertedPendulum 2 $(\times 10^2)$ & $ 0.16 \pm 0.05 $ & $ 5.86 \pm 0.87 $ \\ \hline
\toprule\bottomrule
Walker 1 $(\times 10^2)$ & $ 2.96 \pm 0.16 $ & $ 3.79 \pm 0.34 $ \\ \hline
Walker 2 $(\times 10^2)$ & $ 2.53 \pm 0.22 $ & $ 4.16 \pm 0.34 $ \\ \hline
Walker 3 $(\times 10^2)$ & $ 2.82 \pm 0.17 $ & $ 3.90 \pm 0.14 $ \\
\toprule\bottomrule
Small HalfCheetah 1 $(\times 10^2)$ & $ -0.89 \pm 0.80 $ & $ 3.44 \pm 0.95 $ \\ \hline
Small Hopper 1 $(\times 10^2)$ & $ 4.37 \pm 0.40 $ & $ 4.44 \pm 0.42 $ \\
\toprule
\end{tabular}
}
\end{center}
\end{table}

\section{Learning Curve}\label{apdx:training_curve_all}
\begin{figure}[t]
  \begin{center}
    \includegraphics[trim=0 0 0 0, clip, width=\hsize] {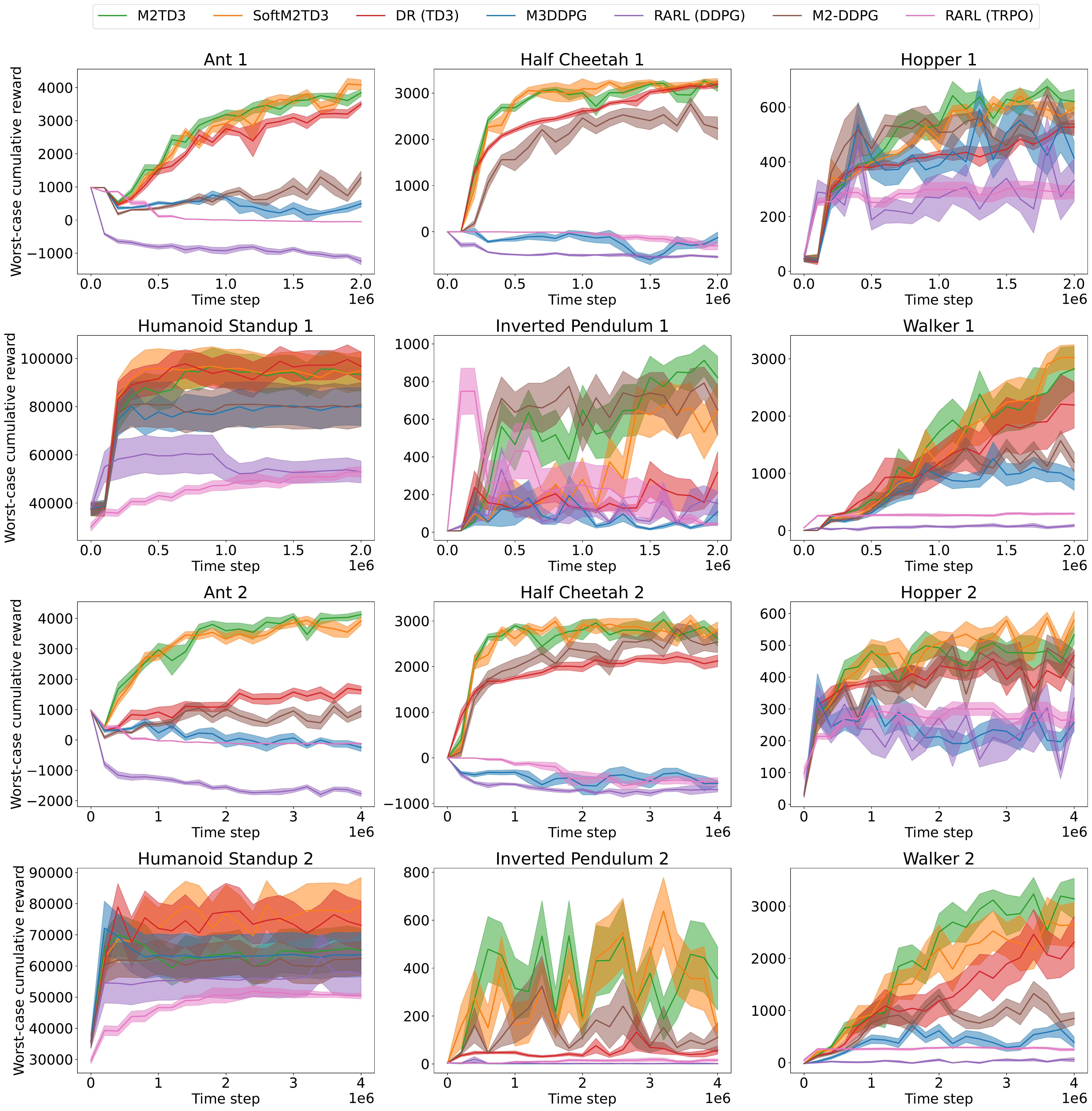}%
    \caption{Learning curve for worst-case performance: the average (solid line) and standard error (band) of the worst-case cumulative rewards $R_\text{worst}(\mu)$
    }
  \label{fig:training_curve_worst_all}
 \end{center}
\end{figure}

\begin{figure}[t]
  \begin{center}
    \includegraphics[trim=0 0 0 0, clip, width=\hsize] {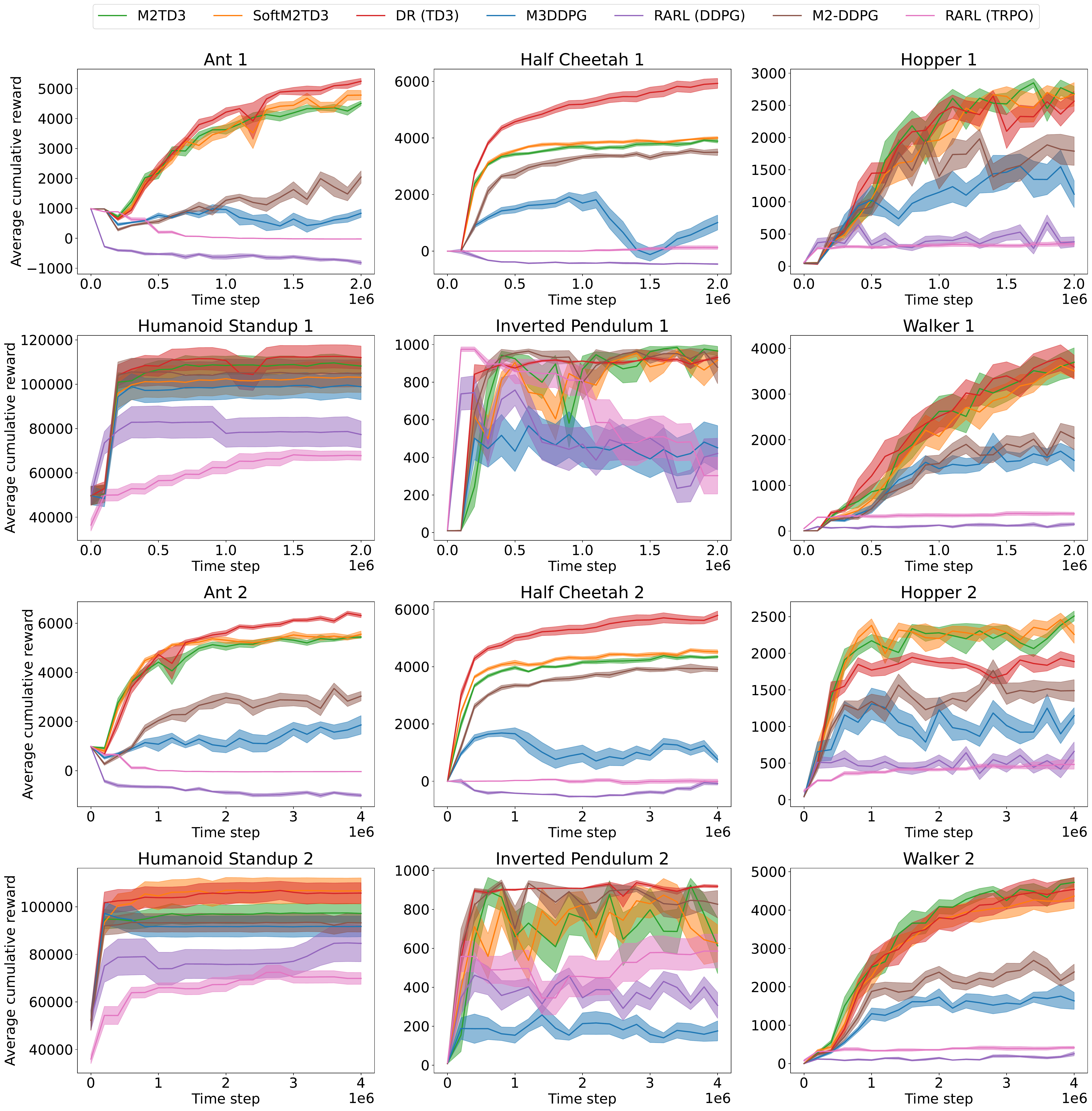}%
    \caption{Learning curve for average performance: the average (solid line) and standard error (band) of the average cumulative rewards $R_\text{average}(\mu)$
    }
  \label{fig:training_curve_average_all}
 \end{center}
\end{figure}

The learning curves for each approach are shown in the \Cref{fig:training_curve_worst_all} and \Cref{fig:training_curve_average_all}.
\Cref{fig:training_curve_worst_all} shows that \proposed{} and SoftM2TD3 tended to perform better in the early stages of learning than DR, even in scenarios where the final worst-case performances of DR, M2TD3, and SoftM2TD3 were competitive.

\section{Cumulative Rewards Under Different Uncertainty Parameters}\label{apdx:each_rewards_1_all}
\begin{figure}[ht]
  \begin{center}
    \includegraphics[trim=0 0 0 0, clip, width=\hsize] {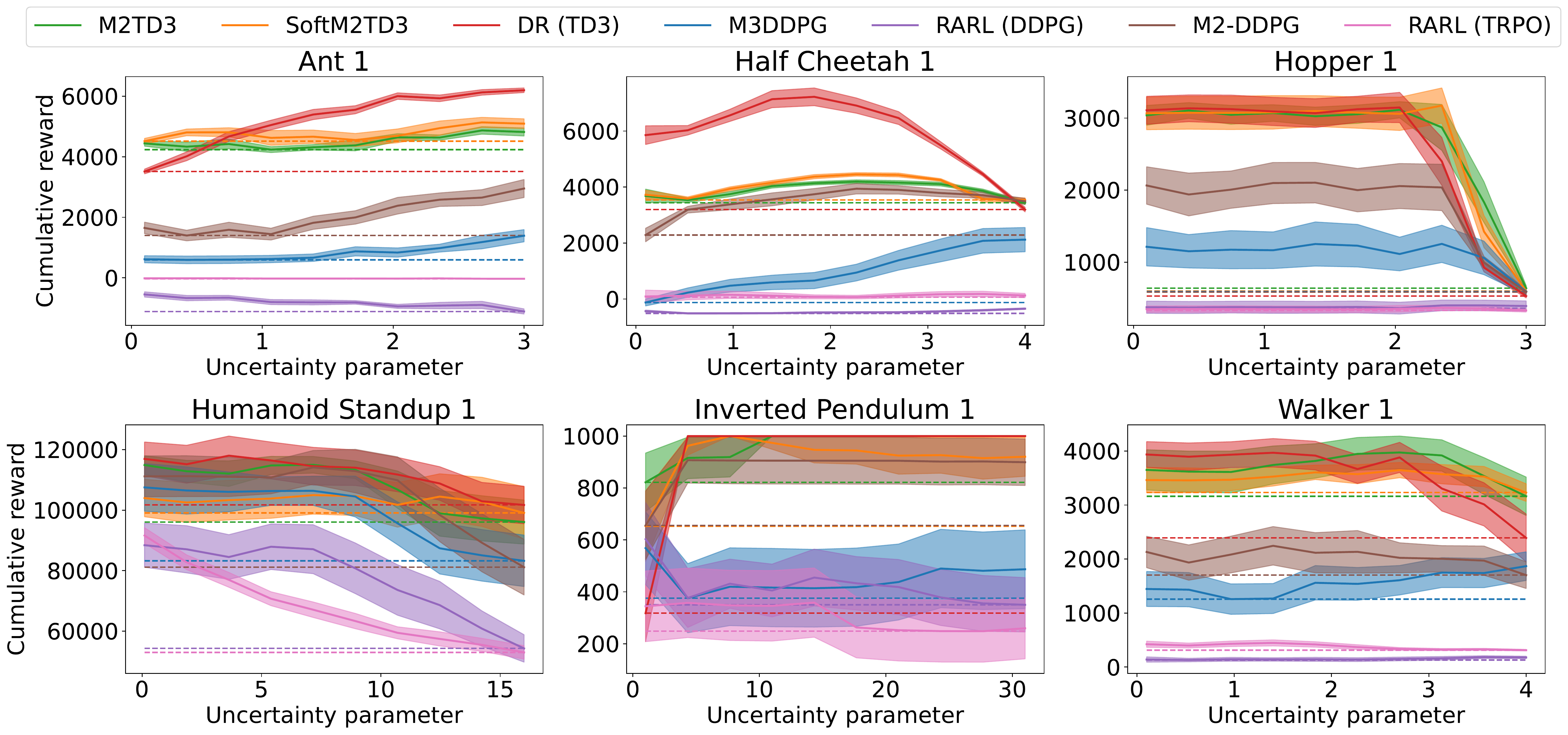}%
    \caption{Cumulative rewards under trained policies for each uncertainty parameter $\omega \in \Omega$. The average (solid line) and standard error (band) for each $\omega \in \Omega$, as well as the worst average value (dashed line) are shown.
    }
  \label{fig:each_rewards_1_all}
 \end{center}
\end{figure}

\begin{figure}[ht]
   \begin{center}
   \begin{minipage}{\hsize}%
     \includegraphics[trim=0 0 0 0, clip, width=\hsize]{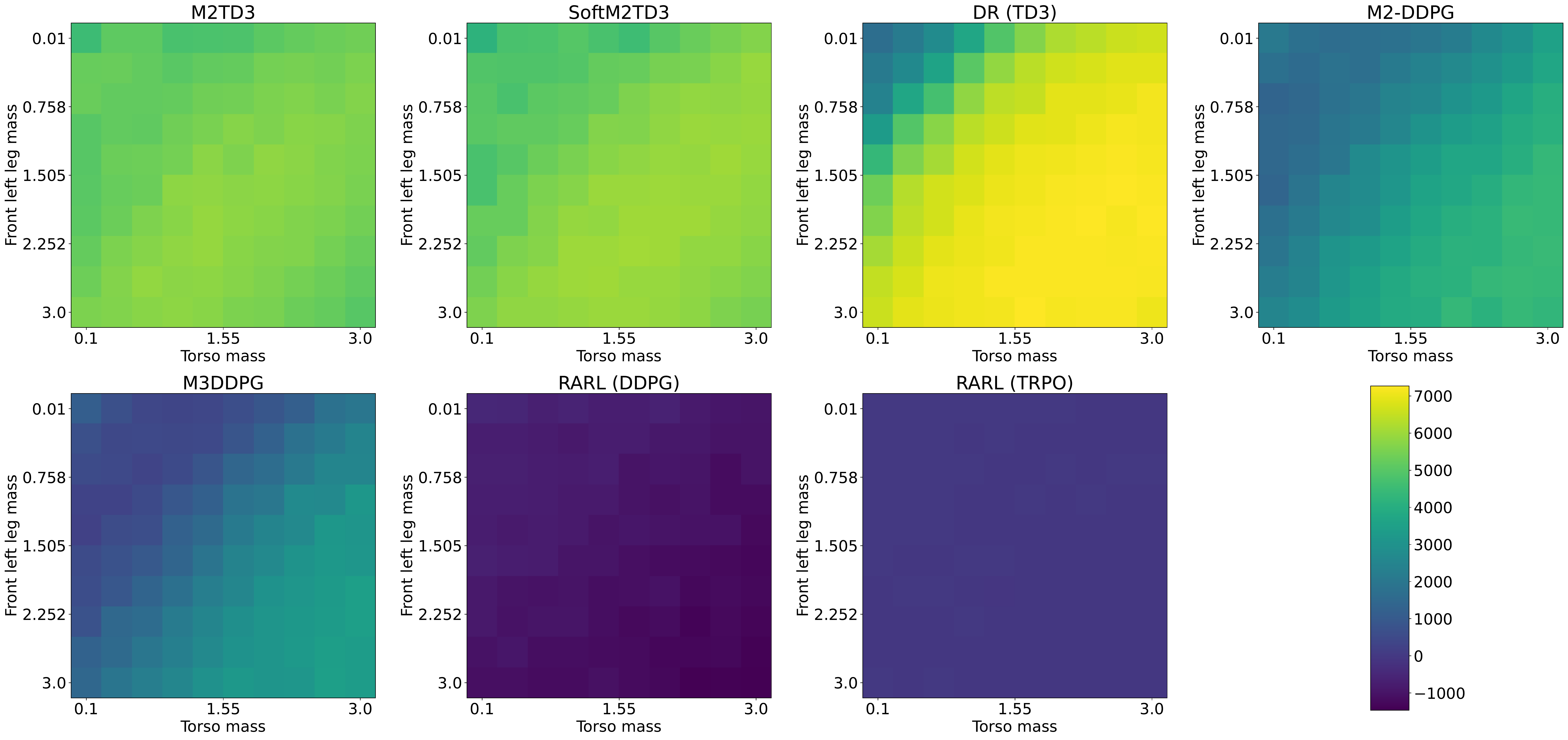}%
     \subcaption{Ant 2}%
   \end{minipage}%
   \\
   \begin{minipage}{\hsize}%
     \includegraphics[trim=0 0 0 0, clip, width=\hsize]{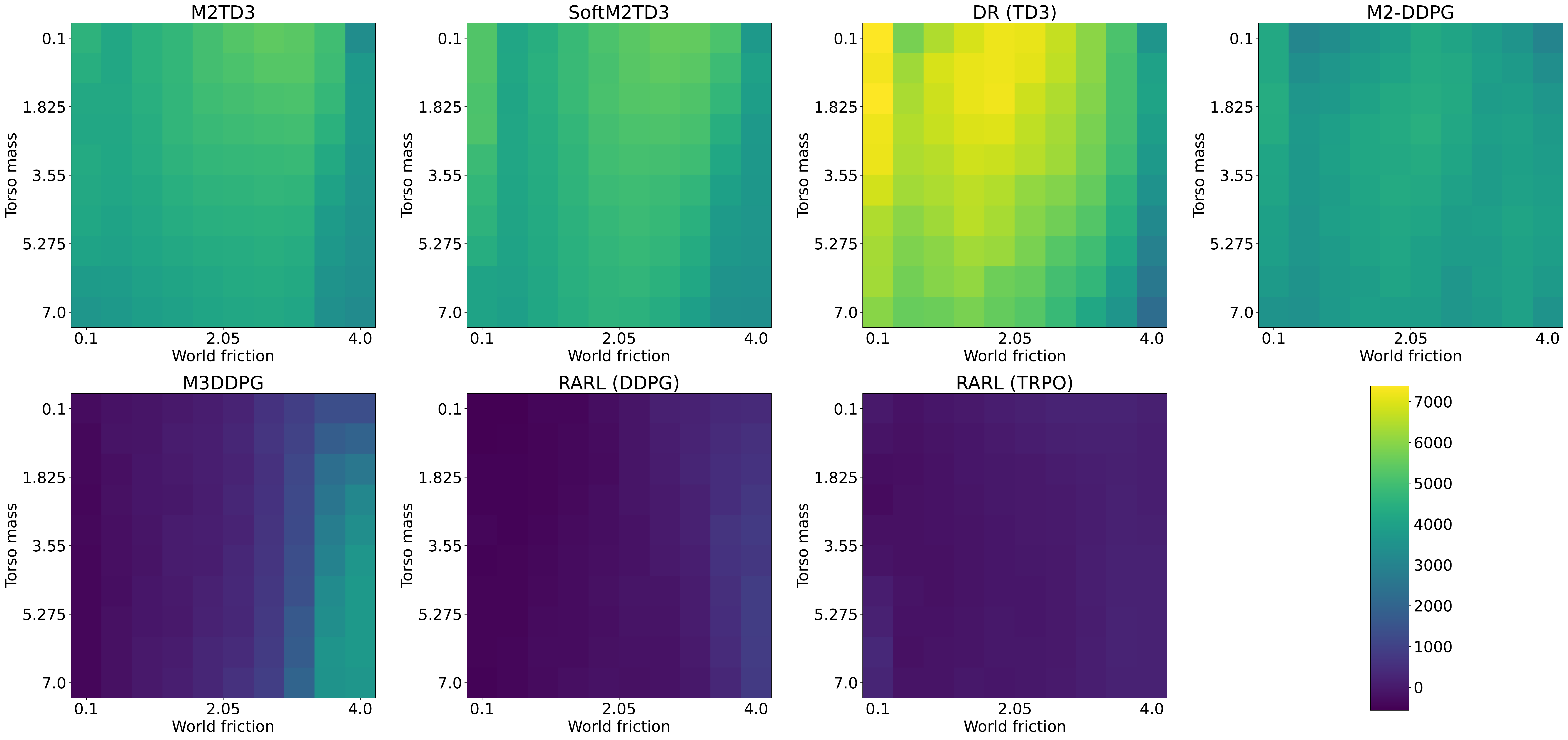}%
     \subcaption{HalfCheetah 2}%
   \end{minipage}%
   \\
   \begin{minipage}{\hsize}%
     \includegraphics[trim=0 0 0 0, clip, width=\hsize]{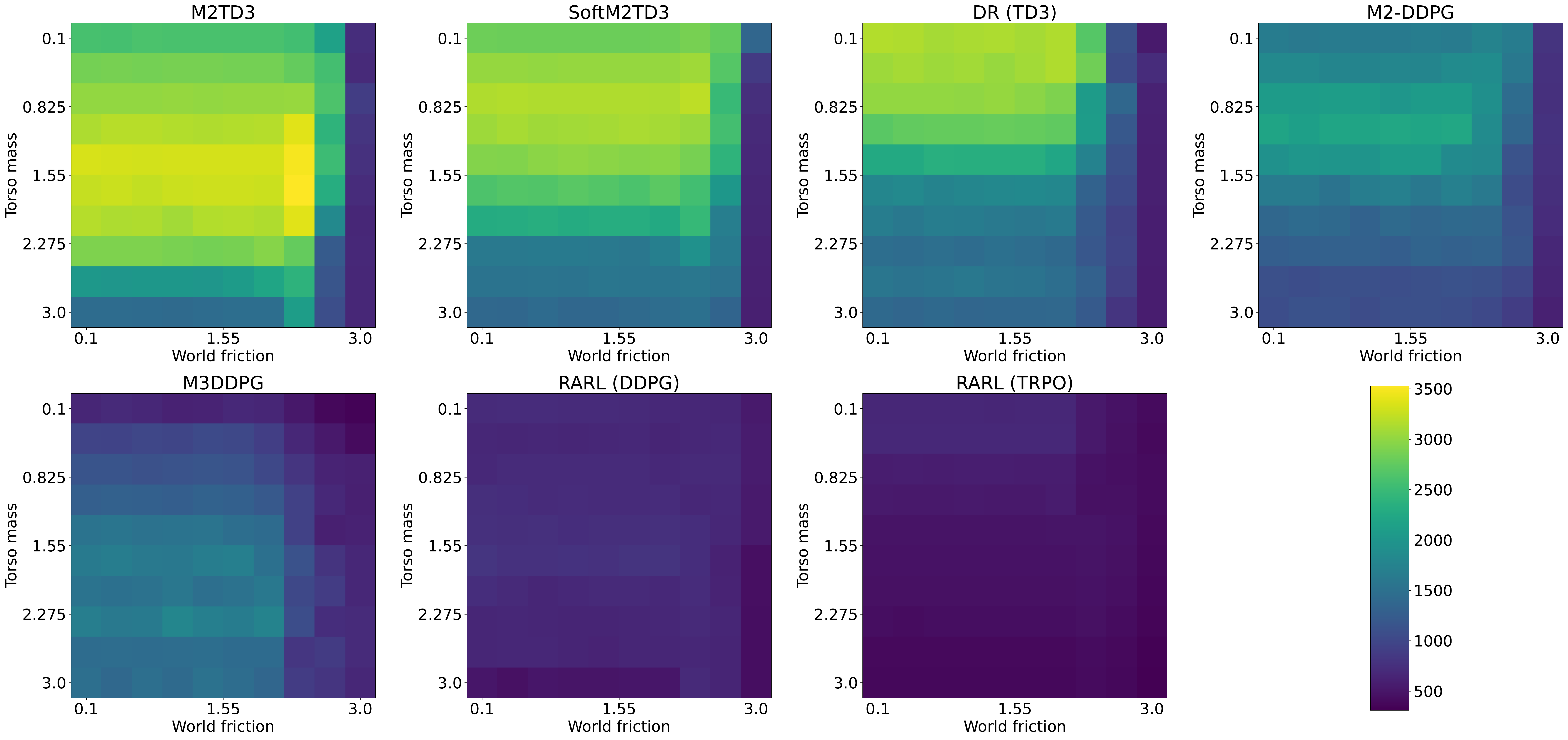}%
     \subcaption{Hopper 2}%
   \end{minipage}%
   \caption{Cumulative rewards under trained policies for each uncertainty parameter $\omega \in \Omega$.}\label{fig: each_rewards_2_all_1}
   \label{fig:each_rewards_2_all_1}
   \end{center}
 \end{figure}

\begin{figure}[ht]
   \begin{center}
      \begin{minipage}{\hsize}%
      \includegraphics[trim=0 0 0 0, clip, width=\hsize]{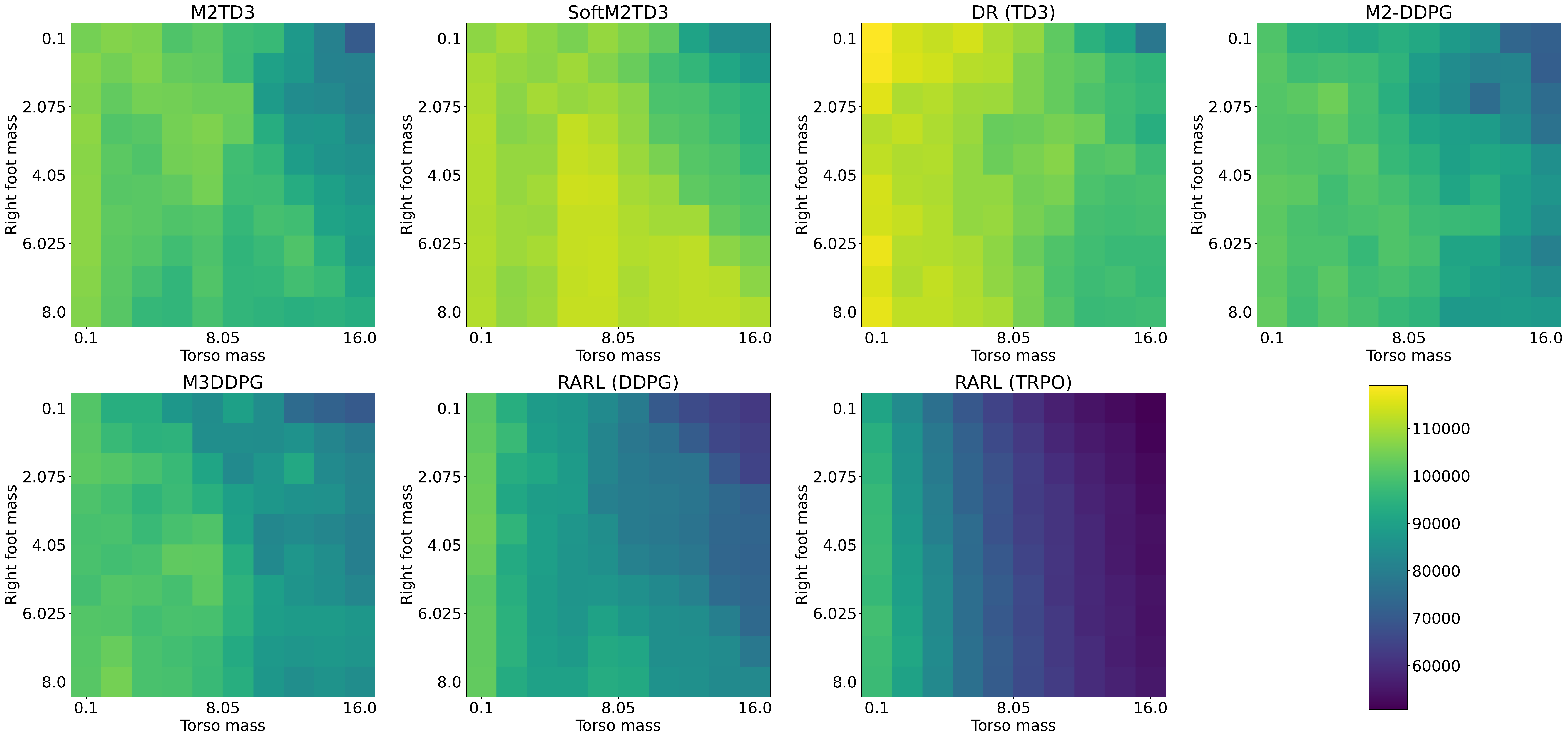}%
      \subcaption{HumanoidStandup 2}%
   \end{minipage}%
   \\
     \begin{minipage}{\hsize}%
     \includegraphics[trim=0 0 0 0, clip, width=\hsize]{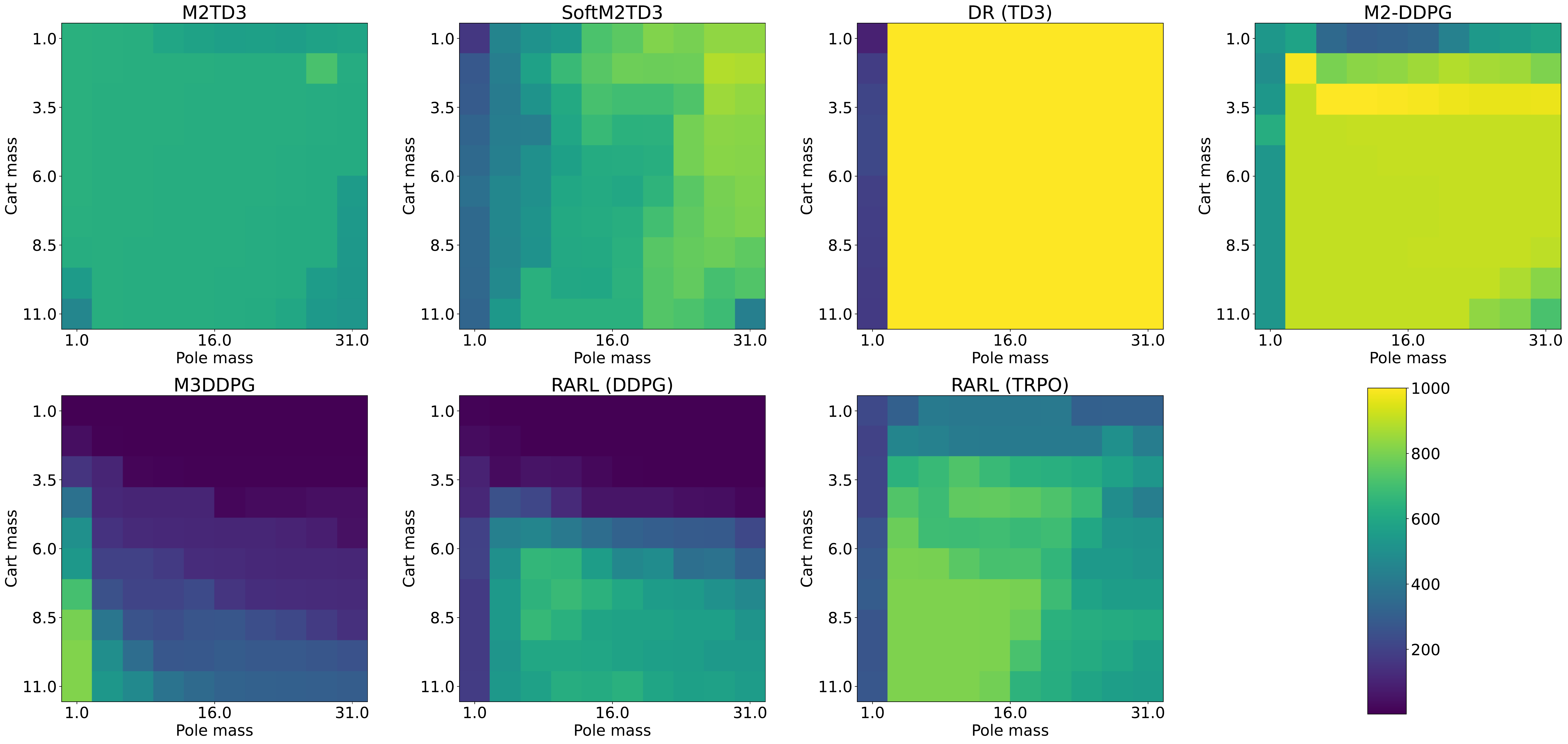}%
     \subcaption{InvertedPendulum 2}%
   \end{minipage}%
   \\
     \begin{minipage}{\hsize}%
     \includegraphics[trim=0 0 0 0, clip, width=\hsize]{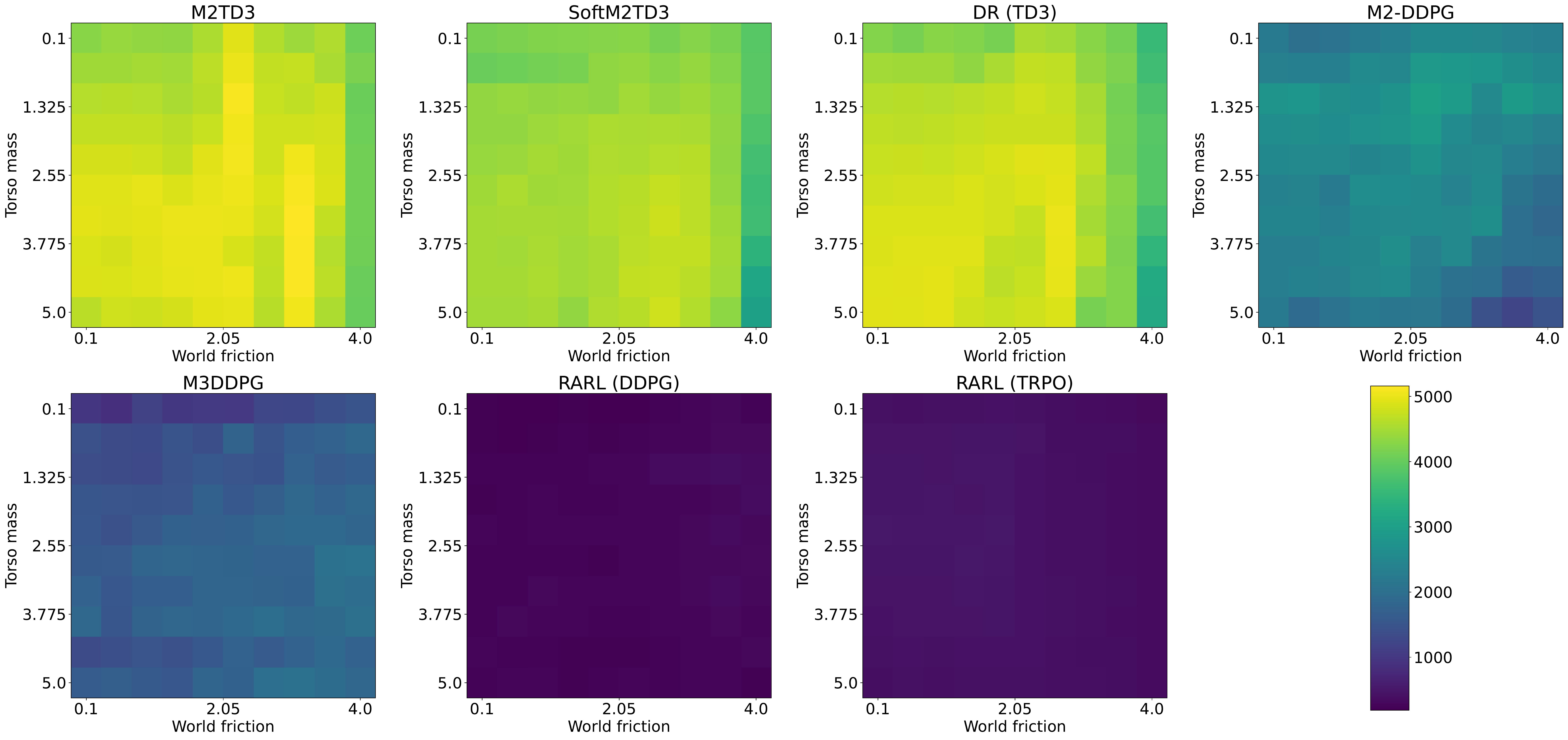}%
     \subcaption{Walker 2}%
   \end{minipage}%
   \caption{Cumulative rewards under trained policies for each uncertainty parameter $\omega \in \Omega$.}\label{fig: each_rewards_2_all_2}
   \label{fig:each_rewards_2_all_2}
   \end{center}
 \end{figure}

\Cref{fig:each_rewards_1_all,fig:each_rewards_2_all_1,fig:each_rewards_2_all_2} show the cumulative rewards of the policies trained by the seven approaches for each $\omega \in \Omega$.
These results show that in many scenarios, DR achieved good performances in a wide range of uncertainty parameters.
However, the differences between the worst-case and best-case performances were relatively large for DR, and the worst-case performances were relatively low.
Unlike DR, M2TD3 exhibited smaller performance differences between the worst-case and the best-case, and its worst-case performances were relatively high.

\section{Multiple Uncertainty Parameter}
\begin{figure}[ht]
  \begin{center}
    \includegraphics[trim=0 0 0 0, clip, width=\hsize] {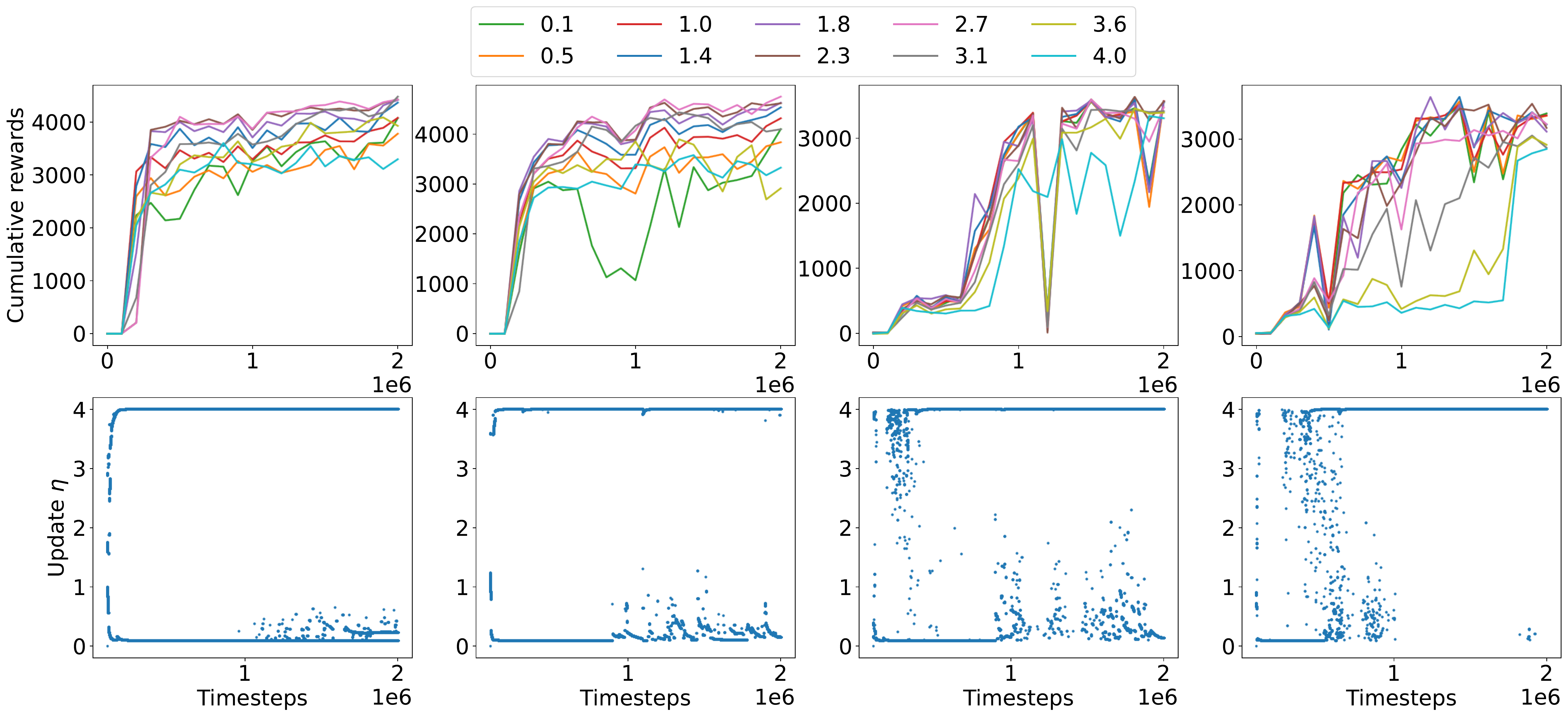}%
    \caption{Cumulative reward under each uncertainty parameter (upper). Uncertainty parameter used to update at each time step (under). From left to right (scenario,  algorithm): (HalfCheetah 1, \proposed), (HalfCheetah 1, SoftM2TD3), (Walker 1, \proposed), (Walker 1, SoftM2TD3).}
  \label{fig:training_curve_each_min_eta}
 \end{center}
\end{figure}

\Cref{fig:training_curve_each_min_eta} shows the learning curves for each uncertainty parameter $\omega \in \Omega$ and the uncertainty parameter $\hat{\omega}$ used to update the actor ($\hat{\omega}_{k'}$ in \Cref{alg:actor}) at each time step during the training.
For each senario, the cumulative rewards under 10 equally spaced uncertainty parameters were evaluated every $1e5$ time steps. Hence, the lowest cumulative reward at each time step is a rough estimate of the worst-case performance.
The uncertainty parameter with the lowest Q-value, $\argmin_{\hat{\omega}_1,\ldots, \hat{\omega}_k, \ldots \hat{\omega}_N} J_t(\theta, \hat{\omega}_k)$, is chosen for the update of the actor among $N$ worst uncertainty parameter candidates. This means that the selected worst-case uncertainty parameter can be different from the ground truth worst-case uncertainty parameter because of incomplete optimization for the worst uncertainty parameter, incomplete training of the critic network, and discrepancy between the cumulative reward and the Q-values.

Focusing on the behavior of \proposed{} on the HalfCheetah 1 senario (left-most side figures),
the cumulative reward (top figure) shows that the uncertainty parameters $0.1$ (green line), $0.5$ (orange line), and $4.0$ (light blue line) alternately came to the bottom during learning, indicating that the worst uncertainty parameter continues to change between these values.
The uncertainty parameter selected during the actor training (bottom figure) were the values around $0.1$ and $4.0$in the early stages of the training, and the values around $0.5$ were used in the middle of the training.
This indicates that the algorithm was able to track the change of the worst-case uncertainty parameter during the training.
This behavior of \proposed{} and SoftM2TD3 is shown on the other tasks.

\section{Evaluation Under Adversarial External Force} \label{apdx:evaluate_under_adv-ex-force}

With a small modification of the proposed approaches, we can apply it to the situation conforming to \cite{DBLP:conf/icml/PintoDSG17}, where the model misspecification is expressed by an external force given by an adversarial agent. In this section, we describe the necessary modification of the proposed approaches and compare the worst-case performance with baselines.

\paragraph{Extension of \proposed{}}
In this setting, we deal with an MDP $\mathcal{M} = \langle S, A_p, A_a, p, p_0, r, \gamma\rangle$, where $A_p$ and $A_{a}$ are the action spaces of the protagonist agent and adversarial agent, respectively, and they are assumed to be continuous.
The transition probability density $p:S \times A_p \times A_a \times S \rightarrow \mathbb{R}$ and immediate reward $r:S \times A_p \times A_a \rightarrow \mathbb{R}$ are not anymore parameterized by $\omega$ but take the action of the adversarial agent as input.
Unless otherwise specified, the other notations are the same as the ones in \Cref{sec:preliminaries}.
The protagonist and adversarial agents interact with the environment $\mathcal{M}$ using stochastic behavior policies $\beta$ and $\alpha$, respectively.
Let $\rho^{\beta}_{\alpha}(s) = \lim_{T \rightarrow \infty} \frac{1}{T} \Sigma_{t=0}^{T-1} \int_{s_0} q^{t}_{\alpha, \beta}(s \mid s_0) p_0(s_0) \mathrm{d}s_0$ represent the stationary distribution of $s$ under $\beta$ and $\alpha$, where
the step-$t$ transition probability density $q^{t}_{\alpha, \beta}$ is defined as $q^{1}_{\alpha, \beta}(s' | s) = \int_{a^p \in A_p} \int_{a^a \in A_a} q(s' | s, a^p, a^a) \beta(a^p | s) \alpha(a^a | s) \mathrm{d}a_a \mathrm{d}a_p$ and $q^{t}_{\alpha, \beta}(s' | s) = \int_{\bar{s} \in S} q^{t-1}_{\alpha, \beta}(\bar{s} | s) q^{1}_{\alpha, \beta}(s' | \bar{s}) \mathrm{d}\bar{s}$.
The joint stationary distribution of $(s, a^p, a^a)$ is defined as $\rho^{\beta}_{\alpha}(s, a^p, a^a) = \beta(a^p | s) \alpha(a^a | s) \rho^{\beta}_{\alpha}(s)$.

We extend the action value function as a function of state $s$, protagonist agent's action $a^p$, and adversarial agent's action $a^a$, namely,
\begin{multline}
  Q^{\mu}_{\xi}(s, a^p, a^a) \\
  = \mathbb{E}[R_t \mid s_t=s, a_{t}^{p} = a^p, a_{t}^{a} = a^a, s_{t+k+1} \sim q(\cdot \mid s_{t+k}, \mu(s_{t+k}), \xi(s_{t+k}))\ \forall k \geq 0] \enspace ,
\end{multline}
where $\mu$ and $\xi$ represent the policies of the protagonist agent and the adversarial agent, respectively, to be trained.
We suppose that they are parameterized by $\theta_p$ and $\theta_a$, respectively.

The objective of \proposed{} is extended as
\begin{equation}
  \max_{\theta_p \in \Theta} \min_{\theta_a \in \Theta} J(\theta_p, \theta_a; \phi^*) \quad \text{s.t.} \quad \phi^* \in \argmin_{\phi \in \Phi} L(\phi; \theta_p, \theta_a) \enspace,
\end{equation}
where the critic loss function $L$ is extended as
\begin{multline}
  L(\phi, \theta_p, \theta_a) := \int_{s \in S} \int_{a^p \in A_p} \int_{a^a \in A_a} (T^{\mu_{\theta_p}}_{\xi_{\theta_a}}[Q_\phi](s, a^p, a^a) - Q_\phi(s, a^p, a^a))^2 \\
  \times \rho^{\beta}_{\alpha}(s, a^p, a^a) \mathrm{d}s \mathrm{d}a^p \mathrm{d}a^a \enspace,
\end{multline}
where $T^{\mu_{\theta_p}}_{\xi_{\theta_a}}$ is a function satisfying
\begin{equation}
  T^{\mu_{\theta_p}}_{\xi_{\theta_a}}[Q](s, a^p, a^a) = r_\omega(s, a^p, a^a) + \gamma \int_{s' \in S} Q(s', \mu_{\theta_p}(s'), \xi_{\theta_a}(s')) q(s'|s, a^p, a^a) \mathrm{d}s' \enspace.
\end{equation}

The max-min objective function $J$ of the actor network is defined as
\begin{equation}
  J(\theta_p, \theta_a; \phi) := \int_{s \in S} Q_{\phi}(s, \mu_{\theta_p}(s), \xi_{\theta_a}(s)) \rho^{\beta}_{\alpha}(s) \mathrm{d}s \enspace ,
\end{equation}
and $\rho^{\beta}_{\alpha}(s, a^p, a^a)$ and $\rho^{\beta}_{\alpha}(s)$ are approximated by the replay buffer $B$ that stores the trajectories obtained by the interaction using $\beta$ and $\alpha$.
Let $\{(s_i, a^{p}_i, a^{a}_i, r_i, s'_i)\}_{i=1}^{M} \subset B$ be mini-batch samples taken uniformly randomly from the replay buffer.
The approximated objective function used for the critic update is
\begin{align}
  \tilde{L}(\phi) &= \frac{1}{M} \sum_{i=1}^{M} (y_i - Q_{\phi}(s_i, a^p_i, a^a_i))^2 ,
\end{align}
where $y_i = r_i+\gamma \cdot Q_{\phi}(s_i', \mu_{\theta_p}(s_i'), \xi_{\theta_a}(s_i'))$.
The approximated objective function used for the actor update is
\begin{equation}
  \tilde{J}(\theta_p, \theta_a) = \frac{1}{M} \sum_{i=1}^M Q_\phi(s_i, \mu_{\theta_p}(s_i), \xi_{\theta_a}(s_i)) \enspace .
\end{equation}
Replacing \eqref{eq:tildel} and \eqref{eq:tildej} with above defined functions, we obtain \proposed{} for this setting.

\paragraph{Experiment}
We used the tasks provided in \cite{DBLP:conf/icml/PintoDSG17}\footnote{\url{https://github.com/lerrel/gym-adv}}.
For each trained policy, the worst-case performance is estimated by fixing the trained policy of the protagonist agent and training the adversarial agent's policy to minimize the protagonist agent's performance.
To evaluate the worst-case performance of each approach, we performed $5$ independent training for $2 \times 10^6$ time steps for each approach.
The trial is indexed as $n \in \{1, \dots, 5\}$.
Then, for each obtained protagonist policy, we trained the adversarial policy for $2 \times 10^6$ time steps.
We performed the adversarial policy training three times, and they are indexed as $m \in \{1, 2, 3\}$.
During the training of the adversarial policy, we recorded the performance of the protagonist agent under the adversarial policy at time step $et \in \{10^5, 2\times 10^{5}, \dots, 2 \times 10^{6}\}$ (every $10^5$ time steps), denoted as $R(\mu, \xi_{m,et})$.
Then, the worst-case performance of a protagonist policy $\mu$ was estimated by $R_\text{worst}(\mu) = \min_{m =1,2,3}\min_{et \in \{10^5, 2\times 10^{5}, \dots, 2 \times 10^{6}\}} R(\mu, \xi_{m,et})$.
The average and standard error of $R_\text{worst}(\mu)$ over $5$ trials are reported for each approach.
The parameters and network architecture of the protagonist agent used in all methods and the adversarial agents used in M2TD3, M2-DDPG, M3DDPG, RARL (TD3), and RARL (TRPO) are the same as in the situation that the uncertainty parameter is directly encoded by $\omega$.
In the situation that the uncertainty parameter was directly encoded by $\omega$, multiple uncertainty parameters were trained, but in this setting, only one adversarial agent was trained.
The adversarial TD3 agents used to estimate worst-case performance were similar to the parameters and network architecture used in \cite{fujimoto2018addressing}.

The result is shown in \Cref{table:rarl_setting_result}.

\proposed{} showed better worst-case performance than DR in all but the HalfCheetahAdv-v1 senarios.
The reference performances were comparable to those of DR.
Compared to those of RARL, \proposed{} showed competitive or superior performances, both in the worst-case and reference-case.
Comparing M2-DDPG (the proposed approach based on DDPG instead of TD3) and M3DDPG, we observed similar performances in many scenarios in both reference-case and worst-case, and M2-DDPG significantly outperformed M3DDPG on HopperAdv-v1.

\begin{table}[t]
\begin{center}
\caption{Avg.\ $\pm$ std.\ error of reference performance (performance under no disturbance) and worst-case performance over five trials for each approach}
\label{table:rarl_setting_result}
\scalebox{0.78}{
\begin{tabular}{|c|c|c|c|c|c|c|c|}
\bottomrule
\multicolumn{1}{|c|}{}           & \multicolumn{1}{c|}{TD3}             & \multicolumn{1}{c|}{DR}              & \multicolumn{1}{l|}{M2TD3}           & \multicolumn{1}{l|}{M2-DDPG}          & \multicolumn{1}{l|}{M3DDPG}          & \multicolumn{1}{l|}{RARL (TD3)}      & RARL (TRPO)      \\
\toprule\bottomrule
\multicolumn{8}{|l|}{MuJoCo Environment: HalfCheetahAdv-v1 ($\times 10^4$)} \\ \hline
\multicolumn{1}{|c|}{reference}  & \multicolumn{1}{c|}{$1.13 \pm 0.09$} & \multicolumn{1}{c|}{$1.21 \pm 0.03$} & \multicolumn{1}{c|}{$1.15 \pm 0.09$} & \multicolumn{1}{c|}{$1.19 \pm 0.04$} & \multicolumn{1}{c|}{$1.21 \pm 0.03$} & \multicolumn{1}{c|}{$1.04 \pm 0.09$} & $0.68 \pm 0.27$  \\ \hline
\multicolumn{1}{|l|}{worst-case} & \multicolumn{1}{l|}{$1.02 \pm 0.11$} & \multicolumn{1}{l|}{$1.12 \pm 0.05$} & \multicolumn{1}{l|}{$1.07 \pm 0.09$} & \multicolumn{1}{l|}{$1.13 \pm 0.06$} & \multicolumn{1}{l|}{$1.13 \pm 0.02$} & \multicolumn{1}{l|}{$1.00 \pm 0.10$} & $0.33 \pm 0.32$  \\
\toprule\bottomrule
\multicolumn{8}{|l|}{MuJoCo Environment: HopperAdv-v1 ($\times 10^3$)} \\ \hline
\multicolumn{1}{|l|}{reference}  & \multicolumn{1}{l|}{$3.48 \pm 0.16$} & \multicolumn{1}{l|}{$3.55 \pm 0.14$} & \multicolumn{1}{l|}{$3.25 \pm 0.56$} & \multicolumn{1}{l|}{$2.63 \pm 0.52$} & \multicolumn{1}{l|}{$1.66 \pm 0.33$} & \multicolumn{1}{l|}{$2.42 \pm 0.94$} & $0.29 \pm 0.07$  \\ \hline
\multicolumn{1}{|l|}{worst-case} & \multicolumn{1}{l|}{$0.50 \pm 0.13$} & \multicolumn{1}{l|}{$1.79 \pm 1.25$} & \multicolumn{1}{l|}{$2.64 \pm 0.84$} & \multicolumn{1}{l|}{$2.16 \pm 0.53$} & \multicolumn{1}{l|}{$0.69 \pm 0.23$} & \multicolumn{1}{l|}{$0.82 \pm 0.45$} & $0.26 \pm 0.08$  \\
\toprule\bottomrule
\multicolumn{8}{|l|}{MuJoCo Environment: InvertedPendulumAdv-v1 ($\times 10^3$)}\\ \hline
\multicolumn{1}{|l|}{reference}  & \multicolumn{1}{l|}{$1.00 \pm 0.00$} & \multicolumn{1}{l|}{$1.00 \pm 0.00$} & \multicolumn{1}{l|}{$1.00 \pm 0.00$} & \multicolumn{1}{l|}{$1.00 \pm 0.00$} & \multicolumn{1}{l|}{$1.00 \pm 0.00$} & \multicolumn{1}{l|}{$0.63 \pm 0.46$} & $0.10 \pm 0.07$  \\ \hline
\multicolumn{1}{|l|}{worst-case} & \multicolumn{1}{l|}{$0.03 \pm 0.01$} & \multicolumn{1}{l|}{$0.23 \pm 0.39$} & \multicolumn{1}{l|}{$0.83 \pm 0.35$} & \multicolumn{1}{l|}{$0.70 \pm 0.31$} & \multicolumn{1}{l|}{$0.80 \pm 0.38$} & \multicolumn{1}{l|}{$0.04 \pm 0.01$} & $0.03 \pm 0.01$  \\
\toprule\bottomrule
\multicolumn{8}{|l|}{MuJoCo Environment: SwimmerAdv-v1 ($\times 10^2$)} \\ \hline
\multicolumn{1}{|l|}{reference}  & \multicolumn{1}{l|}{$1.42 \pm 0.09$} & \multicolumn{1}{l|}{$1.22 \pm 0.25$} & \multicolumn{1}{l|}{$1.38 \pm 0.09$} & \multicolumn{1}{l|}{$1.51 \pm 0.10$} & \multicolumn{1}{l|}{$1.51 \pm 0.06$} & \multicolumn{1}{l|}{$1.21 \pm 0.05$} & $0.21 \pm 0.10$  \\ \hline
\multicolumn{1}{|l|}{worst-case} & \multicolumn{1}{l|}{$0.96 \pm 0.15$} & \multicolumn{1}{l|}{$0.60 \pm 0.48$} & \multicolumn{1}{l|}{$1.14 \pm 0.08$} & \multicolumn{1}{l|}{$1.19 \pm 0.04$} & \multicolumn{1}{l|}{$1.21 \pm 0.04$} & \multicolumn{1}{l|}{$0.80 \pm 0.05$} & $-0.71 \pm 0.37$ \\
\toprule\bottomrule
\multicolumn{8}{|l|}{MuJoCo Environment: Walker2dAdv-v1 ($\times 10^3$)}\\ \hline
\multicolumn{1}{|l|}{reference}  & \multicolumn{1}{l|}{$4.18 \pm 0.44$} & \multicolumn{1}{l|}{$4.27 \pm 0.61$} & \multicolumn{1}{l|}{$4.59 \pm 1.05$} & \multicolumn{1}{l|}{$2.81 \pm 0.81$} & \multicolumn{1}{l|}{$2.26 \pm 0.50$} & \multicolumn{1}{l|}{$3.95 \pm 0.27$} & $0.32 \pm 0.08$  \\ \hline
\multicolumn{1}{|l|}{worst-case} & \multicolumn{1}{l|}{$3.74 \pm 0.59$} & \multicolumn{1}{l|}{$3.61 \pm 0.93$} & \multicolumn{1}{l|}{$4.26 \pm 1.04$} & \multicolumn{1}{l|}{$1.75 \pm 0.62$} & \multicolumn{1}{l|}{$1.63 \pm 0.32$} & \multicolumn{1}{l|}{$3.11 \pm 0.13$} & $0.14 \pm 0.08$  \\
\toprule
\end{tabular}
}\end{center}
\end{table}
\end{document}